\documentclass{article}

\PassOptionsToPackage{numbers, compress}{natbib}
\usepackage[final]{neurips_2024}
\usepackage{placeins}
\usepackage{hyperref}
\usepackage{url}
\usepackage[noend]{algpseudocode}
\usepackage{float}
\usepackage{graphicx}
\usepackage{subcaption}
\usepackage{algorithm}
\usepackage[utf8]{inputenc} 
\usepackage[T1]{fontenc}    
\usepackage{hyperref}       
\usepackage{url}            
\usepackage{booktabs}       
\usepackage{amsfonts}       
\usepackage{nicefrac}       
\usepackage{microtype}      
\usepackage{xcolor}         
\usepackage{amsmath}
\usepackage{lipsum} 
\usepackage{graphicx} 
\usepackage{wrapfig} 
\usepackage{url}
\usepackage{titletoc}
\usepackage{xcolor}

\newtheorem{definition}{Definition}

\newcommand{\ourtitle}{Continual Nonlinear ICA-based Representation Learning}
\newcommand{\mb}{\mathbf}
\newcommand{\bs}{\boldsymbol}
\newcommand{\indep}{\perp \!\!\! \perp}

\newtheorem{lemma}{Lemma}
\newtheorem{remark}{Remark}
\newtheorem{theorem}{Theorem}
\newtheorem{proposition}{Proposition} 

\title{Continual Learning of Nonlinear Independent Representations}

\author{
Boyang Sun$^{1}$, Ignaiver Ng$^{2}$, Guangyi Chen$^{2,1}$, Yifan Shen$^{1}$, Qirong Ho$^{1}$, Kun Zhang$^{2,1}$~\\$^1$  Mohamed bin Zayed University of Artificial Intelligence  \\$^2$ Carnegie Mellon University
}

\begin{document}

\maketitle

\begin{abstract}
Identifying the causal relations between interested variables plays a pivotal role in representation learning as it provides deep insights into the dataset. Identifiability, as the central theme of this approach, 
normally hinges on leveraging data from multiple distributions (intervention, distribution shift, time series, etc). Despite the exciting development in this field, a practical but often overlooked problem is, what if those distribution shifts happen sequentially? In contrast, any intelligence possesses the capacity to abstract and refine learned knowledge sequentially; lifelong learning. In this paper, with a particular focus on the nonlinear independent component analysis (ICA) framework, we move one step forward toward the question of enabling models to learn meaningful (identifiable) representations in a sequential manner, termed continual causal representation learning. 
We theoretically demonstrate that model identifiability progresses from a subspace level to a component-wise level as the number of distributions increases. 
Empirically, We show that our method achieves performance comparable to nonlinear ICA methods trained jointly on multiple offline distributions and, surprisingly the incoming new distribution doesn't necessarily benefit the identification of all latent variables.


\end{abstract}

\section{Introduction}

In many data-driven problems, observations can be considered as the output of mathematical functions applied to the underlying representations. The process of learning those meaningful representations based on the measured variables, known as representation learning, is crucial in modern machine learning fields with applications spanning from visions \citep{wang2020deep, bengio2013representation, doersch2015unsupervised, kolesnikov2019revisiting} to natural language processing \citep{Liu_2020, mohamed2022self, vaswani2017attention, collobert2011natural}. However, traditional approaches to representation learning often fall short by merely capturing statistical correlation without considering the underlying causal structure, which is crucial for achieving robust generalization across different tasks and datasets \citep{schölkopf2021causal}.  Therefore, identifying causal relations between relevant variables is indispensable in representation learning to comprehend the intricate relationships within the datasets.

Discovery of causal relations is never an easy problem. In many scenarios, we want to benefit the identifiability of the causal structure by leveraging non-i.i.d data. In traditional causal discovery, it's well known the causal structure can only be identified up to Markov equivalent class without assuming functional class \citep{Squires_2022} in the i.i.d case. However, interventions, seen as an "active distribution shift", can break the asymmetry and reveal the causal direction. Besides that, a series of studies indicate that distribution shifts and heterogeneous data can improve the identifiability of causal structure \citep{huang2020causal, tian2013causal}.

When the interested variables are not directly measured, causal representation learning (CRL)\citep{schölkopf2021causal} aims at recovering the latent causal variables and their causal structures from the observations. However, similar to causal discovery, learning identifiable representations from i.i.d data is highly challenging. 
In fact, the identifiability of nonlinear independent component analysis (ICA), as the simplest case of CRL where all latents are independent, is proven to be impossible without further assumptions \citep{comon1994independent}. To address this problem, existing work often relies on functional constraints \citep{zheng2022identifiability, lachapelle2022disentanglement, buchholz2022function, zheng2023generalizing} or distributional assumptions \citep{hyvarinen2016unsupervised, hyvarinen2017nonlinear, hyvarinen2018nonlinear, khemakhem2019variational}.

Recent advances in this area aim to extend beyond independent latent variables and recover their causal graph. These advancements often rely on parametric or graph assumptions \citep{ cai2019triad, huang2022latent, xie2020generalized, jin2024structural} regarding the latent causal structure or, commonly, exploit non-i.i.d data. For instance, \citep{vonkügelgen2022selfsupervised} explores identifiability from a counterfactual perspective, while \citep{vonkügelgen2023nonparametric} investigates multi-environment data arising from unknown interventions. Furthermore, \citep{brehmer2022weakly} examines data from distinct paired interventions, and \citep{daunhawer2023identifiability} focuses on multi-modality data. Additionally, \citep{zhang2024causal} proposes a nonparametric general framework.


Despite the exciting developments in this field, we should acknowledge that, whether through intervention, distribution shift, or counterfactual approaches, mathematically, all these methods simply indicate that \textbf{there exist multiple distributions for the causal variables of interest, enabling the learning of meaningful (identifiable) representations}. However, these approaches implicitly assume that we can simultaneously access data from multiple distributions, which is often difficult to achieve in real-world scenarios. In contrast, any intelligent agents are able to continually abstract and refine learned knowledge. This raises a direct and fundamental question: how to enable the model to learn identifiable representations with sequential arriving distributions? This challenge leads us to the concept of \textit{continual causal representation learning (CCRL)}.

In this paper, with a particular focus on the nonlinear ICA framework, we present a novel approach to learning causal representation in continually arriving distributions. Distinct from traditional continual classification tasks, CCRL requires that the model leverages the changes in distribution continually to refine the learned representations, ultimately achieving identifiability. This implies that the problem cannot be segregated into discrete local learning tasks, such as learning causal representation within individual distributions and subsequently fusing them.
In this context, we conduct a theoretical examination of the relationships between model identification and the number of observed distributions. Our research indicates that the identifiability increases with the inclusion of additional distributions. In particular, subspace identification can be achieved with $n+1$ distributions, while component-wise identification necessitates $2n+1$ distributions or more. This indicates that when the distribution count is inadequate ($n+1$), we can only identify the manifold spanned by a subset of latent variables. However, by utilizing the new side information from arriving distributions, we can further disentangle this subset and improve the identifiability.

This discovery motivates us to develop a method that retains prior knowledge and refines it using information derived from incoming distributions, a process reminiscent of human learning mechanisms. To realize CCRL, we employ two objectives: (1) the reconstruction of observations within the current distribution, and (2) the preservation of reconstruction capabilities for preceding distributions via gradient constraints.
To accomplish these goals, we apply Gradient Episodic Memory (GEM) \citep{lopez2017gradient} to constrain the model's gradients. GEM aligns the gradients of the new domain with those of prior distributions by eliminating factors within the current distribution that are detrimental to previous distributions.
Through empirical evaluations, we demonstrate that our continual approach delivers performance on par with nonlinear ICA techniques trained jointly across multiple offline distributions.
We show that the identifiability of the latent causal variables strengthens as the number of observed distributions increases.
Interestingly, our theoretical findings indicate that the new distribution does not necessarily benefit the identification of all latent variables, validated by the experiments.

\section{Related Work}

\textbf{Causal representation learning.} Beyond conventional representation learning, causal 
representation learning aims to identify the underlying causal generation process and recover the latent causal variables. There are pieces of work aiming towards this goal. For example, it has been demonstrated in previous studies that latent variables can be identified in linear-Gaussian models by utilizing the vanishing Tetrad conditions \citep{spearman1928pearson}, as well as the more general concept of t-separation \citep{JMLR:v7:silva06a}. Additionally, the Generalized Independent  Noise (GIN) condition tried to identify a linear non-Gaussian causal graph \citep{xie2020generalized}. However, all of these methods are constrained to the linear case while nonlinear ICA provides a promising framework that learns identifiable latent causal representations based on their non-linear mixture. However, the identifiability of nonlinear ICA has proven to be a challenging task \citep{hyvarinen1999nonlinear}, which always requires further assumptions as auxiliary information, such as temporal structures~\citep{sprekeler2014extension}, non-stationarities~\citep{hyvarinen2016unsupervised, hyvarinen2017nonlinear}, or a general form as auxiliary variable~\citep{hyvarinen2018nonlinear}. These methods indicate that sufficient distributions (changes) are crucial for ensuring the identifiability of nonlinear ICA. In this paper, we consider the scenario that changing distributions may arrive not simultaneously but sequentially or even not adequately. 

\textbf{Continual learning.} In conventional machine learning tasks, the model is trained on a dedicated dataset for a specific task, then tested on a hold-out dataset drawn from the same distribution. However, this assumption may contradict some real-world scenarios, where the data distribution varies over time. It motivates researchers to explore continual learning to enable an artificial intelligence system to learn continuously over time from a stream of data, tasks, or experiences without losing its proficiency in the ones it has already learned. 
The most common setting is class incremental recognition~\citep{rebuffi2017icarl,hou2019learning,van2021class}, where new unseen classification categories with different distributions arrive sequentially. To solve this problem, existing methods are commonly divided into three categories. 
Regulization-based methods~\citep{riemer2018learning,zeng2019continual,farajtabar2020orthogonal,saha2021gradient,tang2021layerwise,wang2021training} add the constraints on the task-wise gradients to prevent the catastrophic forgetting when updating network weights for new arriving distributions.
Memory-based methods~\citep{robins1995catastrophic,rebuffi2017icarl,lopez2017gradient,chaudhry2018efficient,chaudhry2019continual,hu2019overcoming,kemker2017fearnet,shin2017continual,pellegrini2020latent,van2021class}
propose to store previous knowledge in a memory, such as a small set of examples, a part of weights, or episodic gradients to alleviate forgetting.
Distillation-based methods~\citep{li2017learning,rebuffi2017icarl,hou2019learning,castro2018end,wu2019large,yu2020semantic,tao2020topology,liu2020mnemonics,mittal2021essentials} remember the knowledge trained on previous tasks by applying knowledge distillation between previous network and currently trained network.
Please note that CCRL is distinct from conventional class incremental recognition. It is because CCRL needs to leverage the distribution change to identify the latent variables. This implies that the problem cannot be divided into discrete local learning tasks, such as learning representation within observing data from individual distributions and then merging them together, while training separate networks for different tasks will definitely reach state-of-the-art performance in a continual classification learning scenario.
Thus, we introduce a memory model to store the information of previous distributions and use it to adjust the model parameters.

\section{Identifiable Nonlinear ICA with Sequentially Arriving Distributions}


In this section, we conduct a theoretical examination of the relationship between model identification and the number of distributions. Initially, we introduce the causal generation process of our model (in Section~\ref{Sec: problem}), which considers the dynamics of changing distributions. Subsequently, we demonstrate that model identifiability improves with the inclusion of additional distributions. More specifically, we can achieve component-wise identification with $2n+1$ distributions (in Section~\ref{Sec: 2n+1}), and subspace identification with $n+1$ distributions (in Section~\ref{Sec: n+1}). Building on these theoretical insights, we introduce our method for learning independent causal representation in the context of continually emerging distributions (in Section~\ref{Sec: gem}).

\subsection{Problem Setting}
\label{Sec: problem}
\begin{wrapfigure}{r}{0.3\textwidth} 
    \centering
\includegraphics[width=0.32\textwidth]{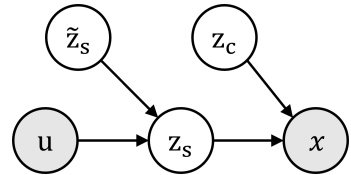} 
    \caption{\textbf{Data generation process.} $\mb{x}$ is influenced by variables $\mb{z}_s$ (change with different distributions $\mb{u}$) and invariant variables $\mb{z}_c$. }
    \label{data gen}
    \vspace{-0.5cm}
\end{wrapfigure}

As shown in Figure~\ref{data gen}, we consider the data generation process as follows: 
\begin{equation}  \label{data generation eq}
    \mb{z}_c \sim p_{\mb{z}_c}, \quad \Tilde{\mb{z}}_s \sim p_{\Tilde{\mb{z}}_s}, \quad \mb{z}_s = f_{\mb{u}}(\Tilde{\mb{z}}_s), \quad \mb{x} = g(\mb{z}_c, \mb{z}_s),
\end{equation}
where $\mb{x} \in \mathcal{X} \subseteq \mathbb{R}^d$ are the observations mixed by latent variables $\mb{z} \in \mathcal{Z} \subseteq \mathbb{R}^n$ through an invertible and smooth nonlinear function $\mb{g}: \mathcal{Z}\rightarrow \mathcal{X}$ ($d\geq n$), which is also called a $\mathcal{C}^2$ diffeomorphism. The latent variables $\mb{z}$ can be partitioned into two groups: changing variables $\mb{z}_s \in \mathcal{Z}_s \subseteq \mathbb{R}^{n_s}$ whose distribution changes across the distribution indicator $\mb{u}$, and invariant variables $\mb{z}_c \in \mathcal{Z}_c \subseteq \mathbb{R}^{n_c}$ whose distribution remains invariant. In this paper, we refer $\mb{u}$ as the \textit{domain}.
Given $T$ distributions in total, we have $p_{\mb{z}_s|\mb{u}_{k}} \not = p_{\mb{z}_s|\mb{u}_{l}} , p_{\mb{z}_c|\mb{u}_{k}} = p_{\mb{z}_c|\mb{u}_{l}}$ for all $k, l \in \{1,\dots,T\}, k\not=l$.
We parameterize the distribution change for changing variables $\mb{z}_s$ as the function of $\mb{u}$ to its parent variables $\Tilde{\mb{z}}_s$, i.e. $\mb{z}_s = f_{\mb{u}}(\Tilde{\mb{z}}_s)$. 
 One can understand this setting with the following example: suppose the higher level variables follow Gaussian distribution, i.e., $\Tilde{\mb{z}}_s \sim \mathcal{N}(\mb{0}, \mb{I})$, and $\mb{u}$ could be a vector denoting the variance of the distribution. The combination of $\mb{u}$ with $\Tilde{\mb{z}}_s$ will produce a Gaussian variable with different variances at different distributions. 

The objective of nonlinear ICA is to recover the latent variables $\mb{z}_s$ and $\mb{z}_c$ given the observation $\mb{x}$ and domain variables $\mb{u}$ by estimating the unmixing function $\mb{g}^{-1}$. In this paper, we consider the case where distributions arrive sequentially, i.e., we aim to recover the latent variables by sequentially observing $\mb{x}|\mb{u}_1, \mb{x}|\mb{u}_2,\dots, \mb{x}|\mb{u}_T$. 

\subsection{Identifiability Theory of Nonlinear ICA}
\label{sec: Ident}


The identifiability is the key to nonlinear ICA to guarantee meaningful recovery of the latent variables. Mathematically,
the identifiability of a model is defined as 
\begin{equation} \label{idn_any_model}
    \forall(\bs{\theta}, \bs{\theta}'): \quad p_{\bs{\theta}}(\mb{x}) = p_{\bs{\theta}'}(\mb{x}) \Longrightarrow \bs{\theta} = \bs{\theta}' ,
\end{equation}
where $\boldsymbol{\theta}$ represents the parameter generating the observation $\mb{x}$.
That is, if any two different choices of model parameter $\boldsymbol{\theta}$ and $\boldsymbol{\theta}'$ lead to the same distribution, then this implies that $\bs{\theta}$ and $\bs{\theta}'$ are equal \citep{khemakhem2019variational}. 
For our data generation defined in \eqref{data gen}, we have $\bs{\theta} = (g, \mb{z}_c, \mb{z}_s )$, and $\bs{\theta}' = (\hat{g}, \hat{\mb{z}}_c, \hat{\mb{z}}_s)$ which denotes the estimated mixing function, estimated invariant variables, and estimated changing variables respectively. Thus, a fully identifiable nonlinear ICA needs to satisfy at least two requirements: the ability to reconstruct the observation and the complete consistency with the true generating process. 
Unfortunately, current research cannot achieve this level of identifiability without further assumptions that are considerably restrictive.
Therefore, existing works typically adopt a weaker notion of identifiability. In the following, we discuss two types of identifiability for the changing variable, and show that the identifiability progressively increases from subspace identifiability to component-wise one by incorporating more distributions.


In this work, we follow \citep{kong2022partial} and assume our estimated latent process $(\hat{g}, \hat{\mb{z}}_c, \hat{\mb{z}}_s)$ could generate  observation $\hat{\mb{x}}$ with identical distribution with observation $\mb{x}$ generated by the true latent process $(g, \mb{z}_c, \mb{z}_s)$, i.e.,
\begin{equation} \label{match_dis}
    p_{\mb{x}|\mb{u}}(\mb{x}'| \mb{u}') = p_{\mb{\hat{x}}|\mb{u}}(\mb{x}'| \mb{u}'), \quad    \mb{x'} \in \mb{\mathcal{X}}, \mb{u'} \in \mb{\mathcal{U}} .
\end{equation}


\subsubsection{Component-wise Identifiability for Changing Variable}
\label{Sec: 2n+1}
First, we show that the changing variable can be identified up to permutation and component-wise invertible transformation with sufficient changing distributions. Specifically, for the true latent changing variable $\mb{z}_s$, there exists an invertible function $h = g ^{-1}\circ \hat{g}: \mathbb{R}^{n_s} \rightarrow \mathbb{R}^{n_s} $ such that $\mb{z}_s = h (\hat{\mb{z}}_s)$, where $h$ is composed of a permutation transformation $\pi$ and a component-wise nonlinear invertible transformation $A$, i.e., $ \hat{g} = g \circ \pi \circ A$ \footnote{\text{More formally "component-wise nonlinear identifiability" as it doesn't require exactly identify each element.}}. That is, the estimated variable $\hat{z}_j$ and the true variable $z_i$ have a one-to-one correspondence with an invertible transformation for $\forall i,j \in \{1,\dots,n_s\}$. We have the following lemma from \citep{kong2022partial}.

\begin{lemma}\label{lemma1}
Suppose that the data generation process follows \eqref{data generation eq} and that the following assumptions hold:
\begin{enumerate}
\item The set $\{\mb{z} \in \mathcal{Z} \mid p(\mb{z}) = 0\}$ has measure zero.
\item The probability density given each domain should be sufficiently smooth. i.e., $p_{\mb{z}|\mb{u}}$ is  at least second-order differentiable.
\item Given domain $\mb{u}$, every element of latent variable $\mb{z}$ should be independent with each other. i.e., $z_i \indep z_j  | \mb{u}$ for $ i,j \in\{1,\dots, n\}$ and $i\not=j $.
\item For any $\mb{z}_s \in \mathcal{Z}_s$, there exists $2n_s + 1$ values of $\mb{u}$, such that for $k=1,\dots,2n_s$, $i=1,\dots,n_s$, the following matrix is invertible:
\[
\hspace{-1.1cm}
\begin{bmatrix}
\phi_1''(\mb{1},\mb{0}) & \hdots & \phi_{i}''(\mb{1},\mb{0}) & \hdots & \phi_{n_s}''(\mb{1},\mb{0})  & \phi_1'(\mb{1},\mb{0}) & \hdots & \phi_{i}'(\mb{1},\mb{0}) & \hdots & \phi_{n_s}'(\mb{1},\mb{0})\\
\vdots & \ddots & \vdots & \vdots  & \vdots  &  \vdots & \vdots & \ddots & \vdots & \vdots\\
\phi_1''(\mb{k}, \mb{0}) & \hdots & \phi_i''(\mb{k}, \mb{0})  & \hdots & \phi_{n_s}''(\mb{k},\mb{0}) &  \phi_1'(\mb{k}, \mb{0}) & \hdots & \phi_i'(\mb{k}, \mb{0})  & \hdots & \phi_{n_s}'(\mb{k},\mb{0}) \\
\vdots & \ddots & \vdots & \vdots  & \vdots  &  \vdots & \vdots & \ddots & \vdots & \vdots\\
\phi_{1}''(\mb{2n_s}, \mb{0}) & \hdots & \phi_i''(\mb{2n_s}, \mb{0})  & \hdots & \phi_{n_s}''(\mb{2n_s},\mb{0}) &  \phi_1'(\mb{2n_s}, \mb{0}) & \hdots & \phi_i'(\mb{2n_s}, \mb{0})  & \hdots & \phi_{n_s}'(\mb{2n_s},\mb{0}) \\
\end{bmatrix},
\]
where
\[
\phi_i''(\mathbf{k}, 0) := \frac{\partial^2 \log(p_{z|u}(z_i | \mathbf{u}_k))}{\partial z_i^2} - \frac{\partial^2 \log(p_{z|u}(z_i | \mathbf{u}_0))}{\partial z_i^2},
\]
\[
\phi_i'(\mathbf{k}, 0) := \frac{\partial \log(p_{z|u}(z_i | \mathbf{u}_k))}{\partial z_i} - \frac{\partial \log(p_{z|u}(z_i | \mathbf{u}_0))}{\partial z_i}.
\]

are defined as the difference between second-order derivative and first-order derivative of log density of $z_i$ between domain $\mb{u_k}$ and domain $\mb{u_0}$ respectively, 
\end{enumerate}
Then, by learning the estimation $\hat{g}, \hat{\mb{z}}_c, \hat{\mb{z}}_s$ to achieve \eqref{match_dis}, $\mb{z}_s$ is component-wise identifiable. \footnote{We only focus on changing variables $\mb{z}_s$ in this paper. One may refer \citep{kong2022partial} for those who are interested in the identifiability of $\mb{z}_c$.}
\end{lemma}

The proof can be found in Appendix \ref{component}. Basically, the theorem states that if the distribution of latent variables is "complex" enough and each domain brings enough changes to those changing variables, those changing variables $\mb{z}_s$ are component-wise identifiable. 



\textbf{Repeated distributions of partially changing variables.}
 The variable $z_i$ is referred to as the changing variable as long as its distribution change happened across $\mb{u_1}$ to $\mb{u}_T$. However, it's important to note that, for different domains $\mb{u}$, the distribution of some changing variable $z_i$ doesn't change necessarily. Specifically, there may exist different domains with the same distribution for partial changing variables:
\begin{equation}
    p_{\mb{z}|u}(z_i|\mb{u_k}) =  p_{\mb{z}|u}(z_i|\mb{u_l}) \quad \exists  k, l \in \{0,\dots, T\}, k\not=l, i \in \{1,\dots,n_s\}.
\end{equation}
One should note that this scenario is quite common in continual learning.
In practical human experience, we frequently encounter novel information that enhances or modifies our existing knowledge base. Often, these updates only alter specific aspects of our understanding, leaving the remainder intact. This raises an intriguing question about model identifiability: Does such partial distribution change impact the invertibility of the matrix in Assumption 4 of Lemma \ref{lemma1}? In response to this issue, we present the following remark, with further details provided in Appendix \ref{appB}.


\begin{remark} \label{remark1}
    For $n_s \geq 2$ and we use $|S_i|$ to denote the cardinality of non-repetitive distributions of latent changing variable $z_i$ ($1\leq|S_i|\leq T$). If Lemma~\ref{lemma1} hold, then $|S_i| \geq 3$ for every $i \in \{1,\dots,n_s\}$.
\end{remark}



\subsubsection{Subspace Identifiability for Changing Variable}
\label{Sec: n+1}

Although component-wise identifiability is powerful and attractive, holding $2n_s+1$ different distributions with sufficient changes remains a rather strong condition and may be hard to meet in practice. In this regard, we investigate the problem of what will happen if we have fewer distributions. We first introduce a notion of identifiability that is weaker compared to the component-wise identifiability discussed in the previous section.

\begin{definition}[Subspace Identifiability of Changing Variable]
We say that the true changing variables $\mathbf{z}_s$ are subspace identifiable if, for the
estimated changing variables $\hat{\mathbf{z}}_s$ and each changing variable $z_{s,i}$, there exists a function $h_i: \mathbb{R}^{n_s} \rightarrow \mathbb{R}$ such that $z_{s,i} = h_i(\hat{\mb{z}}_s) $.
\end{definition}

We now provide the following identifiability result that uses a noticeably weaker condition (compared to Lemma \ref{lemma1}) to achieve the subspace identifiability defined above, using only $n_s+1$ distributions. 

\begin{theorem} \label{thm1} 
Suppose that the data generation process follows \eqref{data generation eq} and that Assumptions 1, 2, and 3 of Lemma~\ref{lemma1} hold.
For any $\mb{z}_s \in  \mathcal{Z}_s$, we further assume that there exists $n_s + 1$ values of $\mb{u}$ such that for $i = 1,\dots, n_s$ and $k=1,\dots, n_s$,  the following matrix
\[
\begin{bmatrix}
\phi_1'(\mb{1},\mb{0}) & \hdots & \phi_{i}'(\mb{1},\mb{0}) & \hdots & \phi_{n_s}'(\mb{1},\mb{0})\\
\vdots & \ddots & \vdots & \vdots  & \vdots  \\
 \phi_1'(\mb{k},\mb{0}) & \hdots & \phi_i'(\mb{k}, \mb{0})  & \hdots & \phi_{n_s}'(\mb{k},\mb{0}) \\
   \vdots & \vdots & \vdots &  \ddots  & \vdots\\
\phi_1'(\mb{n_s}, \mb{0}) & \hdots & \phi_i'(\mb{n_s}, \mb{0})  & \hdots & \phi_{n_s}'(\mb{n_s},\mb{0}) \\
\end{bmatrix}
\]
is invertible, where
\[\phi'_i(\mb{k},\mb{0}):= \frac{\partial \log(p_{\mb{z}|\mb{u}}(z_i| \mb{u_k}))}{ \partial z_i } - \frac{\partial \log(p_{\mb{z}|\mb{u}}(z_i| \mb{u_0}))}{ \partial z_i }\]
is the difference of first-order derivative of log density of $z_i$ between domain $\mb{u_k}$ and domain $\mb{u_0}$ respectively. Then, by learning the estimation $\hat{g}, \hat{\mb{z}}_c, \hat{\mb{z}}_s$ to achieve  \eqref{match_dis}, $\mb{z}_s$ is subspace identifiable.
\end{theorem}

The proof can be found in Appendix~\ref{subspace}. Basically, Theorem~\ref{thm1} proposes a weaker form of identifiability with relaxed conditions. With $n_s + 1$ different distributions, each true changing variable can be expressed as a function of all estimated changing variables. This indicates that the estimated changing variables capture all information for the true changing variables, and thus disentangle changing and invariant variables. It is imperative to emphasize that, within our framework, the subspace identifiability of changing variables can lead to block-wise identifiability ~\citep{kong2022partial, vonkügelgen2022selfsupervised}. We provide detailed proof of this in Appendix~\ref{subspace}. 
Moreover, it is worth noting that if there is only one changing variable, such subspace identifiability can lead to component-wise level. When integrated with the continual learning scenario, we uncover the following interesting properties.

\begin{wrapfigure}{r}{0.3\textwidth} 
    \centering
    \includegraphics[width=0.3\textwidth]{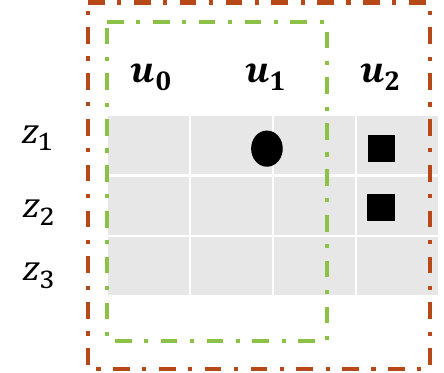} 
    \caption{\textbf{A toy example with three variables and three distributions.} $z_1$ changes in $\mb{u}_1, \mb{u}_2$, $z_2$ changes in $\mb{u}_2$}
    \label{4domain}
\end{wrapfigure}

\textbf{New distributions may impair original identifiability of partial changing variables.}
Consider a toy case where there are three variables with three distributions in total as shown in Figure~\ref{4domain}. The first variable $z_1$ changes in domain $\mb{u}_1$ and both $z_1$ and $z_2$ change in domain $\mb{u}_2$. When considering only distributions $\mb{u}_0$, $\mb{u}_1$,  $z_1$ can achieve subspace identifiability according to Theorem~{\ref{thm1}}.

It is imperative to recognize that, due to the absence of variability in the remaining variables within these distributions, this subspace identifiability inherently aligns with component-wise identifiability, i.e., $z_1$ is component-wise identifiable when only observing $\mb{u}_0, \mb{u}_1$.
However, when considering distributions $\mb{u_0}, \mb{u_1}, \mb{u_2}$, the component-wise identifiability for $z_1$ can't be guaranteed anymore, and instead, we can only promise subspace identifiability for both $z_1$ and $z_2$. In this case, information from domain $\mb{u}_2$ can be viewed as "noise" for $z_1$. 
Contrasted with the traditional joint learning setting, where the data of all distributions are overwhelmed, the continual learning setting offers a unique advantage. It allows for achieving and maintaining original identifiability, effectively insulating it from the potential "noise" introduced by newly arriving distributions. 
In Section~\ref{experiment}, we empirically demonstrate that the causal representation of $z_1$ obtained through continual learning exhibits better identifiability compared to that obtained through joint training.  
In addition, another straightforward property is discussed in the Appendix \ref{another property}.

\subsection{Method}
\label{Sec: gem}
In this section, we leverage the insight of the identifiability theory from previous section to develop our estimation method.

\begin{figure}
    \centering
    \includegraphics[width=0.9\linewidth]{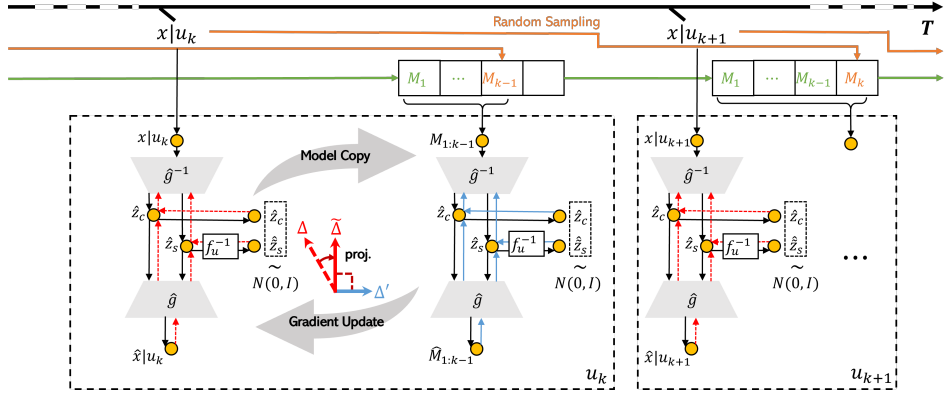}
    \caption{\textbf{ Overall framework}. For the data from new domain $\mb{x}|\mb{u}_i$, we calculate the gradients $\Delta$ and $\Delta'$ of our model with both current data and previous memory. Then, we project the gradient $\Delta$ to $\Tilde{\Delta}$ using Equation \ref{qp} when the angle between $\Delta$ and $\Delta'$ is larger than 90 degrees. Finally, we randomly sample a part of the data in the current domain and add them to the memory bank.}  
    \label{diagram}
    \vspace{-0.3cm}
\end{figure}

\textbf{Generative model.} 
As shown in Lemma~\ref{lemma1} and Theorem~\ref{thm1}, we are aiming at estimating causal process $\hat{g}, \hat{\mb{z}}_c, \hat{\mb{z}}_s$ to reconstruct the distribution of observation. As shown in Figure~\ref{diagram}, we construct a Variational Autoencoder (VAE) with its encoder $q_{\hat{g}^{-1}_\mu, \hat{g}^{-1}_\Sigma}(\hat{\mb{z}}|\mb{x})$ to simulate the mixing process and the decoder $\hat{g}$ to reconstruct a matched distribution $\hat{\mb{x}} = \hat{g}(\hat{\mb{z}})$. Besides, as introduced in data generation in Equation~\ref{data generation eq}, the changing latent variable is generated as the function of high-level invariance $ \hat{\Tilde{\mb{z}}}_s$ with a specific domain influence $\mb{u}$. Assuming the function is invertible, we employ a flow model to obtain the high-level variable $\hat{\Tilde{\mb{z}}}_s$ by inverting the function, i.e., $\hat{\Tilde{\mb{z}}}_s =  \hat{f}^{-1}_{\mb{u}} (\mb{\hat{z}}_s)$.
To train this model, we apply an ELBO loss as:
\begin{equation} \label{vaeloss}
\begin{aligned}
    \mathcal{L}(\hat{g}^{-1}_\mu, \hat{g}^{-1}_\Sigma, \hat{f}_{\mb{u}}, \hat{g}) = \mathbb{E}_{\mb{x}}\mathbb{E}_{\mb{\hat{z}} \sim q_{\hat{g}^{-1}_\mu, \hat{g}^{-1}_\Sigma}} \frac{1}{2}\|x-\hat{x}\|^2 + \alpha KL(q_{\hat{g}^{-1}_\mu, \hat{g}^{-1}_\Sigma}(\hat{\mb{z}}_c| \mb{x}) \|p (\mb{z}_c )  ) \\
    + \beta KL(q_{\hat{g}^{-1}_\mu, \hat{g}^{-1}_\Sigma, \hat{f_\mb{u}}}( \hat{\Tilde{\mb{z}}}_s | \mb{x} \|p (\Tilde{\mb{z}}_s ) ) ,
\end{aligned}
\end{equation}
where $\alpha$ and $\beta$ are hyperparameters controlling the factor as introduced in \citep{higgins2017betavae}. To make the \eqref{vaeloss} tractable, we choose the prior distributions $p (\Tilde{\mb{z}}_s )$ and $p (\mb{z}_c )$ as standard Gaussian $\mathcal{N}(\mb{0}, \mb{I})$.

\textbf{Continual causal representation learning.} 
The subspace identifiability theory in Section~\ref{Sec: n+1} implies that the ground-truth solution lies on a manifold that can be further constrained with more side information, up to the solution with component-wise identifiability. Consequently, it is intuitive to expect that when we observe distributions sequentially, the solution space should progressively narrow down in a reasonable manner.

It motivates us to first learn a local solution with existing distributions and further improve it to align with the new arriving domain without destroying the original capacity.  
Specifically, to realize causal representation learning, we employ two objectives: (1) the reconstruction of observations within the current domain, and (2) the preservation of reconstruction capabilities for preceding distributions. In terms of implementation, this implies that the movement of network parameters learning a new domain should not result in an increased loss for the previous distributions.

 To achieve this goal, we found the classical technique GEM \citep{lopez2017gradient} enables constraining the gradient update of network training to memorize knowledge from previous distributions.  The basic intuition of the algorithm can be illustrated with the following toy example: suppose data from those two distributions are denoted as $\{\mathbf{x |u_1}, \mathbf{x |u_2} \}$ and the parameter of the network $\bs{\theta}$ and the loss calculated on data from $k$th domain is denoted as $l(\bs{\theta}, \mb{x | u_k })$.  At the moment of finishing the learning of the first domain, if we don't make any constraints, the model should start the training using data from the second domain with the direction $\frac{\partial l(\bs{\theta}, \mb{x | u_2 }) }{\partial \bs{\theta}}$.

At this moment, if the direction $\frac{\partial l(\bs{\theta}, \mb{x | u_2 }) }{\partial \bs{\theta}}$ happens to have the property
that $\langle \frac{\partial l(\bs{\theta}, \mb{x | u_2 }) }{\partial \bs{\theta}}, \frac{\partial l(\bs{\theta}, \mb{x | u_1 }) }{\partial \bs{\theta}}\rangle  >0 $, the current direction will contribute to both distributions and we remain the direction. Once the $\langle \frac{\partial l(\bs{\theta}, \mb{x | u_2 }) }{\partial \bs{\theta}}, \frac{\partial l(\bs{\theta}, \mb{x | u_1 }) }{\partial \bs{\theta}}\rangle < 0 $ happens, we project the $\frac{\partial l(\bs{\theta}, \mb{x | u_2 }) }{\partial \bs{\theta}}$ to the direction where $\langle \frac{\partial l(\bs{\theta}, \mb{x | u_2 }) }{\partial \bs{\theta}}, \frac{\partial l(\bs{\theta}, \mb{x | u_1 }) }{\partial \bs{\theta}}\rangle = 0 $, the orthogonal direction to $\frac{\partial l(\bs{\theta}, \mb{x | u_1 }) }{\partial \bs{\theta}}$ where no loss increment for previous distributions.
However, there are infinite possible directions satisfying the orthogonal direction requirement. e.g., we can always use the vector containing all zeros. To make the projected gradient as close as the original gradient,  we  solve for the projected gradient $\frac{\partial l(\bs{\theta}, \mb{x | u_2 })}{\partial \bs{\theta}}'$ that minimizes the objective function
\begin{equation}
    \left\| \frac{\partial l(\bs{\theta}, \mb{x | u_2 }) }{\partial \bs{\theta}} - \frac{\partial l(\bs{\theta}, \mb{x | u_2 }) }{\partial \bs{\theta}}' \right\| ^2 \quad \operatorname{s.t.} \quad \frac{\partial l(\bs{\theta}, \mb{x | u_2 }) }{\partial \bs{\theta}}^T \frac{\partial l(\bs{\theta}, \mb{x | u_1 }) }{\partial \bs{\theta}}' \geq 0 .
\end{equation}

Extend into the general case for multiple distributions, we consider the following quadratic programming problem w.r.t. vector $\mb{v}'$:
\begin{equation} \label{qp}
 \min_{\mb{v}'}\| \mb{v} - \mb{v}'\|^2 \quad  \operatorname{s.t.} \quad \mb{B}\mb{v}'   \geq 0 ,
\end{equation}
where $\mb{v}$ denotes the original gradient, $\mb{v}'$ denotes the projected gradient, $\mb{B}$ is the matrix storing all gradients of past distributions. 
Note that practically, we only store a small portion of data for each domain, thus $\mb{B}_i$ is the row of $\mb{B}$ storing the memory gradient $\frac{\partial l(\bs{\theta}, \mb{M | u_i }) }{\partial \bs{\theta}}$, where $\mb{M|u_i} \in \mb{x|u_i}$. We provide the complete procedure in Appendix~\ref{alg}.

\begin{figure*}[ht]
    \centering
\includegraphics[width=0.99\textwidth]{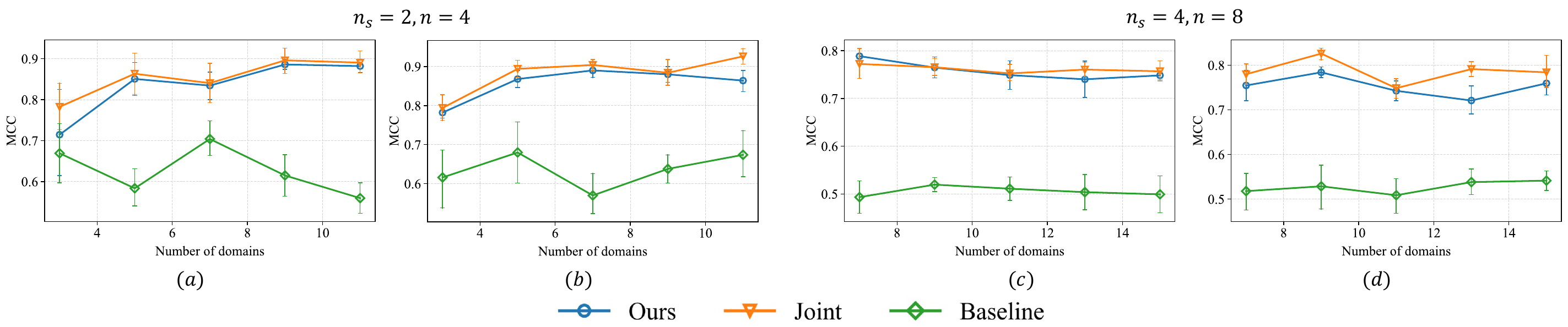}
    \caption{Comparison of MCC for all four datasets with the number of distributions from $2n_s-1$ to $2n_s + 7$. $(a) (c)$ corresponds to Gaussian and $(b) (d)$ corresponds to mixed Gaussian. In this instance, the number of training and the number of testing distributions are equated, which differs from the investigation for increasing distributions.}
     \label{fig:compare}
\end{figure*}

\section{Experiments} \label{experiment}
In this section, we present the implementing details of our method, the experimental results, and the corresponding analysis.

\subsection{Experiment Setup}
\textbf{Data.} 
We follow the standard practice from previous work \citep{hyvarinen2018nonlinear, kong2022partial} and compare our method to the baselines on synthetic data. We generate the latent variables $\mb{z}_s$ for non-stationary and mixed Gaussian distributions with domain-influenced variance and mean, while $\mb{z}_c$ follows standard Gaussian and mixed Gaussian with constant mean and variance. The mixing function is a 2-layer MLP with Leaky-Relu activation. More details are in Appendix \ref{experiment design}.

\textbf{Evaluation metrics.}
We use Mean Correlation Coefficient (MCC) to measure the identifiability of the changing variable $\mb{z}_s$. However, as the identifiability result can only guarantee component-wise identifiability, it may not be fair to directly use MCC between $\hat{\mb{z}}_s$ and $\mb{z}_s$ (e.g. if $\hat{\mb{z}}_s = \mb{z}_s^3$, we will get a distorted MCC value). We thus separate the test data into the training part and test part, and further train separate MLP to learn a simple regression for each $\hat{\mb{z}}_s$  to $\mb{z}_s$ to remove its nonlinearity on the training part and compute the final MCC on the test part. 
We repeat our experiments over 5 or 3 random seeds for different settings.

\subsection{Experimental Results}

\textbf{Comparison to baseline and joint training.}   We evaluate the efficacy of our proposed approach by comparing it against the same model trained on sequentially arriving distributions and multiple distributions simultaneously, referred to as the baseline and theoretical upper bound by the continual learning community. We employ identical network architectures for all three models and examine four distinct datasets, with respective parameters of $\mb{z}_s $ being Gaussian and mixed Gaussian with $n_s=4$, $n=8$, as well as $n_s=2$, $n=4$. Increasing numbers of distributions are assessed for each dataset. Figure~\ref{fig:compare} shows our method reaches comparable performance with joint training. Further visualization can be found in Appendix \ref{vis}.

        
        


\begin{figure}
    \centering
    \includegraphics[width=\linewidth]{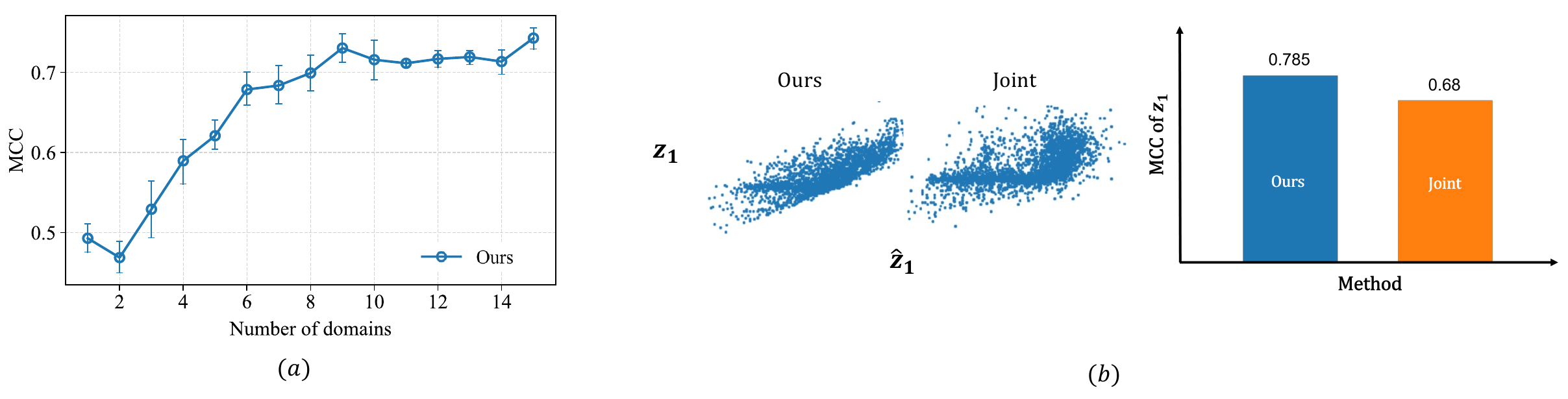}
    \caption{(a) MCC for increasing distributions with models tested on all 15 distributions after training of each domain. (b) Comparison of identifiability for $z_1$ using Joint training and our method qualitatively and quantitatively.      }
    \label{fig:15domain_noniid}
\end{figure}

\textbf{Increasing distributions.} For dataset $n_s=4, n=8$ of Gaussian, we save every trained model after each domain and evaluate their MCC. Specifically, we evaluated the models on the original test dataset, which encompasses data from all 15 distributions.  As shown in part (a) of Figure~\ref{fig:15domain_noniid}, remarkably, increasing distributions lead to greater identifiability results, which align with our expectations that sequential learning uncovers the true underlying causal variables as more information is revealed.
Specifically, we observe that the MCC reaches a performance plateau at 9 distributions and the extra distributions(from 9 to 15) don't provide further improvement. This appears to be consistent with the identifiability theory that $2n_s+1=9$ distributions are needed for identifiability.





\textbf{Discussion: is joint training always better than learning sequentially? Not necessarily.} As discussed in Section~\ref{Sec: n+1}, the new domain may impair the identifiability of partial variables. While joint training always shuffles the data and doesn't care about the order information, learning sequentially to some extent mitigates the impairment of identifiability. 

To test our hypothesis, 
specifically, we conducted an experiment in which both $z_1$ and $z_2$ are Gaussian variables. The variance and mean of $z_1$ change in the second domain, while the other variable changes in the third domain. We then compare our method with joint training only for latent variable $z_1$.  We repeat our experiments with 3 random seeds and the experiment shows that the MCC of our method for $z_1$ reaches up to 0.785 while joint training retains at 0.68 as shown in Figure~\ref{fig:15domain_noniid}(b). In terms of visual contrast, the scatter plot obtained using our method on the left of Figure~\ref{fig:15domain_noniid}(b) exhibits a significantly stronger linear correlation compared to joint training. Further discussions and propositions are provided in Appendix~\ref{dicussion_partial}

\begin{wraptable}{r}{0.5\textwidth}
  \centering
  \vspace{-0.3cm}
  \begin{tabular}{ccccc}
    & T=3  & T=5  & T=7 & T=9  \\
    \hline
    $\hat{n}_s=2$ & $0.782$ & $\mb{0.868}$ & $\mb{0.890}$ & $\mb{0.880}$ \\
    $\hat{n}_s=3$ & $0.781$ & $0.835$ & $0.836$ & $0.834$ \\
    $\hat{n}_s=4$ & $\mb{0.830}$ & $0.861$ & $0.838$ & $0.868$ \\
  \end{tabular}
  \caption{MCC comparison for different preset value of the number of changing variables $\hat{n}_s$}  \label{tb1}
\end{wraptable}

\textbf{Ablation study on prior knowledge of changing variables.} A major limitation of our approach is the requirement for prior knowledge of the number of changing variables. Developing a method to automatically determine the number of changing variables is nontrivial in the continual learning scenario. Therefore, we turn to conducting an ablation study to investigate the sensitivity of this prior knowledge. We implemented the ablation study on mixed Gaussian case with $n=4, n_s=2$ with different five seeds. The results as shown in \text{Table~\ref{tb1}} indicate that our method exhibits relative stability, with a discernible performance decline observed when there is a mismatch between the actual and estimated numbers.






\section{Conclusion}
In this paper, we present a novel approach for the fundamental, but overlooked problem: as long as the learning of identifiable representations relies on multiple distributions, how can we  facilitate this learning in a continual manner?
We believe this approach is unaviodable in practical scenario and just akin to human learning. With a particular focus on nonlinear ICA framework, we examined the relationship between model identification and the number of observed distributions. Our findings indicate that as additional distributions are incorporated, the identifiability of changing variables escalates, with subspace identification achievable with $n_s+1$ distributions and component-wise identification requiring $2n_s+1$ distributions or more. Besides, we briefly show that the introduction of new distributions does not necessarily contribute to all variables. 
Empirical evaluations have demonstrated that our approach achieves performance on par with nonlinear ICA techniques trained jointly across multiple offline distributions, exhibiting greater identifiability with increasing distributions observed.

\vspace{-0.3cm}
\paragraph{Limitation} A obvious limitation of this approach is the requirement for prior knowledge of the number of changing latent variable. Besides that, the gradient based method pose a challenge to the scaling of the algorithm. 
\vspace{-0.3cm}
\paragraph{Future work} 
We believe this paper is only a preliminary exploration of CCRL—there are many issues that need to be addressed. Firstly, this paper focuses solely on the nonlinear Independent Component Analysis (ICA) framework, which is the simplest form of CRL. Recovering the latent variables and their causal structure by observing sequentially arriving distributions is a direct challenge. Secondly, even in causal discovery, how to leverage sequential distribution shifts to benefit the identification of causal relations is worth investigating. Thirdly, it is important to note that humans learn actively and selectively. Correspondingly, understanding how to "smartly" utilize important domains also merits in-depth exploration.

\bibliographystyle{plain}
\bibliography{ref}

\appendix

\newpage
\textit{\large Appendix for}\\ \ \\
  {\large \bf ``\ourtitle''}\
\vspace{.1cm}

\newcommand{\beginsupplement}{%
	\setcounter{table}{0}
	\renewcommand{\thetable}{A\arabic{table}}%
	\setcounter{figure}{0}
	\renewcommand{\thefigure}{A\arabic{figure}}%
	\setcounter{algorithm}{0}
	\renewcommand{\thealgorithm}{A\arabic{algorithm}}%
	\setcounter{section}{0}
	\renewcommand{\thesection}{A\arabic{section}}%
}

\newcommand\DoToC{%
  \startcontents
  \printcontents{}{1}{\hskip10pt\hrulefill\vskip1pt}
 \hskip3pt \vskip1pt \noindent\hrulefill
  }



\beginsupplement

{\large Appendix organization:}

\DoToC 

\section{Proof and Discussion}
We divide our proof into the following parts. First, we start from the matched distribution of \text{the} estimated observation and the real observation, then we show the \text{the} true latent variables can be expressed as invertible transformations of \text{the }estimated variables.
We then use derivatives to construct component-wise relations between \text{the estimated variables with the true latents}. We finally show, with enough domains, we can construct the matrix whose invertibility will force the changing variables subspace identifiable with $n_s+1$ domains and component-wise identifiable with $2n_s+1$ domains.

We start from the matched distribution as introduced in Equation~3: for $ \mb{u'} \in \mb{\mathcal{U}}$
\begin{equation}
        p_{\mb{x}|\mb{u}} = p_{\mb{\hat{x}}|\mb{u}} .
\end{equation}
will imply 
\begin{equation}
    p_{g(\mb{z})|\mb{u}} = p_{\hat{g}(\mb{\hat{z}})|\mb{u} .
}\end{equation}
according to the function of transformation, we can get
\begin{equation}
    p_{g^{-1} \circ g(\mb{z)|\mb{u}}} |J_{g}^{-1}| = p_{g^{-1} \circ \hat{g}(\hat{\mb{z}})|\mb{u}} |J_{g}^{-1}| .
\end{equation}
Let $h := g^{-1} \circ \hat{g}$ to express the transformation from estimated latent variables to real latent variables, i.e., $\mb{z} = h(\mb{\hat{z}})$. As long as both $\hat{g}$ and $g$ are invertible, the transformation \text{$h$} should also be invertible.  We can then get the following
\begin{equation} \label{z=hz}
    p_{\mb{z|u}} = p_{ h (\hat{\mb{z}})|\mb{u} } {.}
\end{equation}
according to the conditional independence(\text{assumption~3}) and nonzero densities(\text{assumption~1}) in Lemma~1, the log density of each marginal distribution can be expressed as 
\begin{equation}
\begin{aligned}
    \log  p_{\mb{z|u }}(\mb{z}) &= \sum_{i=1}^n  \log p_{z_i|\mb{u}}(z_i) ; \\
    \log  p_{\mb{\hat{z}|u }}(\mb{z}) &= \sum_{i=1}^n  \log p_{\hat{z}_i|\mb{u}}(\hat{z}_i) .
\end{aligned}   
\end{equation}
Thus, from Equation~\ref{z=hz} and
according to the function of transformation
\begin{equation}
    p_{\mb{z|u }} = p_{(\hat{\mb{z}})|\mb{u}} |J_{{h}^{-1}}| .
\end{equation}
Take log density on both sides,
\begin{equation}
    \sum_{i=1}^n  \log p_{z_i|\mb{u}}(z_i) =  \sum_{i=1}^n  \log p_{\hat{z}_i|\mb{u}}(\hat{z}_i) + \log |J_{{h}^{-1}}| .
\end{equation}
\text{Simplify the notation as $q_i(z_i, \mb{u}) = \log p_{z_i|\mb{u}}(z_i), \hat{q}_i(\hat{z}_i, \mb{u}) = \log p_{\hat{z}_i|\mb{u}}(\hat{z}_i)$}, the above equation is 
\begin{equation} \label{log_joc}
    \sum_{i=1}^n q_i(z_i, \mb{u}) =  \sum_{i=1}^n \hat{q}_i(\hat{z}_i, \mb{u}) + \log |J_{{h}^{-1}}| . 
\end{equation}
From Equation~\ref{log_joc}, we can see
\begin{equation} \label{A9}
    \sum_{i=1}^n q_i(z_i, \mb{u}) + \log|J_h| = \sum_{i=1}^n \hat{q}_i (\hat{z}_i, \mb{u})
\end{equation}

Until now, we have constructed the relationship between all true latent variables and all estimated variables. In the following sections, we will show how to use the technique of derivatives to establish component-wise relationships between them and how to utilize multi-domain information to eliminate the intractable Jacobian term.

In Section~\ref{subspace}, we show the proof of Theorem~1 and the Proposition~1 inspired by it. In Section~\ref{component}, we show the proof of Lemma~1.  In Section~\ref{appB}, we discuss the case where there are repeated distributions across different domains for partial changing variables and show if there are two or more changing variables, at least three non-repetitive distributions are required for each variable. 

\subsection{Subspace identifiability with $n_{s}+1$ domains} \label{subspace}

Take the derivative of Equation~\ref{A9} with estimated invariant variable $\hat{z}_j$ where \text{$j \in \{n_{s+1}, \dots, n \}$}. We can get 
\begin{equation}
    \sum_{i=1}^n \frac{\partial q_i(z_i, \mb{u})}{\partial z_i} \frac{\partial z_i}{\partial \hat{z}_j} + 
    \frac{\partial \log |J_h|}{\partial \hat{z}_j} = \frac{\partial \hat{q}_j (\hat{z}_j, \mb{u})}{\partial \hat{z}_j}
\end{equation}
The equation allows us to construct the component-wise relation between true latent variable $\mb{z}$ with estimated invariant variables $\hat{\mb{z}}_c$ \text{as expressed using $\frac{\partial z_i}{\partial \hat{z}_j}$}. However, the Jacobian term $\frac{\partial \log |J_h|}{\partial \hat{z}_j}$ is intractable as we have no knowledge about $h$(once we have, everything is solved). 
If we have multiple domains $\mb{u} = \mb{u}_\mb{0}, \dots ,  \mb{u}_\mb{n_s}$, we have $n_s+1$ equations like equation above. We can remove the intractable Jacobian by taking the difference for every equation $\mb{u} = \mb{u_1}, \dots , \mb{u}_\mb{n_s}$ with the equation where $\mb{u} = \mb{u_0}$:
\begin{equation}
    \sum_{i=1}^n (\frac{\partial q_i(z_i, \mb{u_q})}{\partial z_i} - \frac{\partial q_i(z_i, \mb{u_0})}{\partial z_i} ) \frac{\partial z_i} {\partial \hat{z}_j} 
    = 
    \frac{\partial \hat{q}_j (\hat{z}_j, \mb{u_q})}{\partial \hat{z}_j} - 
    \frac{\partial \hat{q}_j (\hat{z}_j, \mb{u_0})}{\partial \hat{z}_j}
\end{equation}
\text{As long as the $j \in \{n_{s+1}, \dots, n \}$}, the distribution of estimated variable $\hat{z}_j$ doesn't change across all domains. \text{The right-hand side of the equation above will be zero}.  Thus,
\begin{equation}
    \sum_{i=1}^n (\frac{\partial q_i(z_i, \mb{u_k})}{\partial z_i} - \frac{\partial q_i(z_i, \mb{u_0})}{\partial z_i} ) \frac{\partial z_i} {\partial \hat{z}_j} 
    = 0
\end{equation}
Similarly, $q_i(z_i,\mb{u})$ remains the same for \text{$i \in \{n_{s+1}, \dots, n\}$}
\begin{equation}
    \sum_{i=1}^{n_s} (\frac{\partial q_i(z_i, \mb{u_k})}{\partial z_i} - \frac{\partial q_i(z_i, \mb{u_0})}{\partial z_i} ) \frac{\partial z_i} {\partial \hat{z}_j} 
    = 0
\end{equation}

Thus, we can have the linear system:
\begin{equation}
    \begin{bmatrix}
        \frac{\partial q_1(z_1, \mb{u_1})}{\partial z_1} -         \frac{\partial q_1(z_1, \mb{u_0})}{\partial z_1} & \hdots &         \frac{\partial q_{n_s}(z_{n_s}, \mb{u_1})}{\partial z_{n_s}} - 
        \frac{\partial q_{n_s}(z_{n_s}, \mb{u_0})}{\partial z_{n_s}} \\
        \vdots & \hdots & \vdots \\
        
        \frac{\partial q_1(z_1, \mb{u}_\mb{n_s})}{\partial z_1} -         \frac{\partial q_1(z_1, \mb{u_0})}{\partial z_1} & \hdots &         \frac{\partial q_{n_s}(z_{n_s}, \mb{u}_\mb{n_s})}{\partial z_{n_s}} - 
        \frac{\partial q_{n_s}(z_{n_s}, \mb{u_0})}{\partial z_{n_s}}        
    \end{bmatrix}
    \begin{bmatrix}
        \frac{\partial z_1}{\partial \hat{z}_j} \\
        \vdots \\
        \frac{\partial z_{n_s}}{\partial \hat{z}_j}
    \end{bmatrix} = 0
\end{equation}
If the matrix above is invertible, its null space will only contain all zeros. Thus, 
$\frac{\partial z_i}{\partial \hat{z}_j} = 0$ for any $i \in \{1,\dots, n_s\}, j \in \{n_{s+1}, \dots, n\}$. That is, $\frac{\partial \mb{z}_s}{\partial \hat{\mb{z}}_c}=0$.
Simplify the notation and define

\begin{equation}
        \phi'_i(\mb{k}):= \frac{\partial \log(p_{\mb{z}|\mb{u}}(z_i| \mb{u_k}))}{ \partial z_i } - \frac{\partial \log(p_{\mb{z}|\mb{u}}(z_i| \mb{u_0}))}{ \partial z_i }
\end{equation}
If the matrix 
\begin{equation}
\begin{bmatrix}
\phi_1'(\mb{1}) & \hdots & \phi_{i}'(\mb{1}) & \hdots & \phi_{n_s}'(\mb{1})\\
\vdots & \ddots & \vdots & \vdots  & \vdots  \\
 \phi_1'(\mb{k}) & \hdots & \phi_i'(\mb{k})  & \hdots & \phi_{n_s}'(\mb{k}) \\
   \vdots & \vdots & \vdots &  \ddots  & \vdots\\
\phi_1'(\mb{n_s}) & \hdots & \phi_i'(\mb{n_s})  & \hdots & \phi_{n_s}'(\mb{n_s}) \\
\end{bmatrix}
\
\end{equation}
is invertible, we can get $\frac{\partial \mb{z}_s}{\partial \hat{\mb{z}}_c}=0$.

 We further look back into the Jacobian matrix which captures the relation true latent variables $\mb{z}$ with the estimated variables $\mb{z}$:
\begin{equation}
J_h = 
    \begin{bmatrix}
\frac{\partial \mb{z}_c}{ \partial \hat{\mb{z}}_c} & \frac{\partial \mb{z}_c}{\partial \hat{\mb{z}}_s} \\
\frac{\partial \mb{z}_s}{\partial \hat{\mb{z}}_c} &
\frac{\partial \mb{z}_s}{\partial \hat{\mb{z}}_s}    
    \end{bmatrix}.
\end{equation}

As long as the transformation $h$ is invertible, the Jacobian matrix $J_h$ should be full rank. Thus, 
The $\frac{\partial \mb{z}_s}{\partial \hat{\mb{z}}_c}$ means that the bottom row of Jacobian above can only contain non-zero in $
\frac{\partial \mb{z}_s}{\partial \hat{\mb{z}}_s}$ . That is, for each true changing variable $z_{s,i}$, it can be written as the function $h_i$ of the estimated changing variables $\hat{\mb{z}}_s$ such that $z_{s,i} = h_i(\hat{\mb{z}}_s)$, which accomplishes the proof.

\begin{proposition}
If Theorem~\ref{thm1} holds, for the estimated changing variables $\hat{\mb{z}}_s$ and true changing variables $\mb{{z}}_s$, there exist an invertible function $h_s: \mathcal{R}^{n_s} \rightarrow \mathcal{R}^{n_s}$ such that $\mb{z}_s = \hat{\mb{z}}_s$ (\textbf{block-wise identifiability}).
\end{proposition}

\textit{Proof} ~We follow the result that $\frac{\partial \mb{z}_s}{\partial \hat{\mb{z}}_c} = 0$ and recall the large Jocabian matrix $J_h$ is invertible, according to the property of invertible block matrix that

\begin{equation}
    \det \begin{bmatrix}
A & B \\
0 & D \\
\end{bmatrix} = \det(A) \det(D).
\end{equation}
Thus, we can derive the determinant of $J_h$ is 
\begin{equation}
    \det(J_h) = \det(\frac{\partial \mb{z}_c}{ \partial \hat{\mb{z}}_c}) \det(\frac{\partial \mb{z}_s}{\partial \hat{\mb{z}}_s}).
\end{equation}
As long as $\det(J_h) \not= 0$ ($J_h$ is full rank), neither $\det(\frac{\partial \mb{z}_c}{ \partial \hat{\mb{z}}_c})$ nor $\det(\frac{\partial \mb{z}_s}{\partial \hat{\mb{z}}_s})$ should equal to $0$. Thus, the transformation from $\mb{z}_s$ to $\hat{\mb{z}}_s$ should be invertible, which accomplishes the proof.

\begin{remark}
We know that $\frac{\partial \mb{z}_s}{\partial \hat{\mb{z}}_c} = 0$, which is kind of trivial intutively as $z_s$ is changing while $\hat{z}_c$ remains the same distribution. However, as the Jacobian matrix $J_h$ is invertible, we can utilize its property that the inverse of a block matrix is 
\begin{equation}
    \begin{bmatrix}
A & B \\
C & D \\
\end{bmatrix}^{-1}
=
\begin{bmatrix}
(A-BD^{-1}C)^{-1} & -(A-BD^{-1}C)^{-1}BD^{-1} \\
-D^{-1}C(A-BD^{-1}C)^{-1} & D^{-1} + D^{-1}C(A-BD^{-1}C)^{-1}BD^{-1} \\
\end{bmatrix}
\end{equation}
Thus, for the inverse of the Jacobian matrix above 
\begin{equation}
    {J_h}^{-1} = 
        \begin{bmatrix}
\frac{\partial \hat{\mb{z}}_c}{ \partial \mb{z}_c} & \frac{\partial \hat{\mb{z}}_c}{\partial \mb{z}_s} \\
\frac{\partial \hat{\mb{z}}_s}{\partial \mb{z}_c} &
\frac{\partial \hat{\mb{z}}_s}{\partial \mb{z}_s}    
    \end{bmatrix}
\end{equation}
The bottom left term $\frac{\partial \hat{\mb{z}}_s}{\partial \mb{z}_c}$ must be zero. This provides more valuable insight, stating that the estimated changing variables cannot be expressed as the function of true invariant variables.
\end{remark}

\vspace{1cm}

\subsection{Component-wise identifiability for $2n_s + 1$ domains} \label{component}

 Differentiating both sides of Equation~\ref{A9} with respect to $\hat{z}_j$, $j \in \{1,\dots, n\}$, we can get
\begin{equation}
    \frac{\partial \hat{q}_j(\hat{z}_j, \mb{u})}{\partial \hat{z}_j} = \sum_{i=1}^n \frac{\partial q_i(z_i, \mb{u})}{\partial z_i} \frac{\partial z_i}{ \partial \hat{z}_j} + \frac{\partial \log |J_{h}|}{\partial \hat{z}_j} .
\end{equation}
Further differentiate with respect to $\hat{z}_q$, $q \in \{1,\dots, n\}, q\not=j$, according to the chain rule,
\begin{equation} \label{nothing}
    0 = \sum_{i=1}^n \frac{\partial^2 q_i(z_i, \mb{u})}{\partial z_i^2} \frac{\partial z_i }{\partial \hat{z}_j} \frac{\partial z_i}{\partial \hat{z}_q } + \frac{\partial q_i(z_i ,\mb{u})}{\partial z_i} \frac{\partial ^2 z_i}{\partial \hat{z}_j \partial \hat{{z}_q}} + \frac{\partial^2 \log|J_{{h}}| }{\partial \hat{z}_j \partial \hat{{z}}_q} . 
\end{equation}
This equation allows us to have the component-wise relation between $\mb{\hat{z}}$ with $\mb{z}$. Following the same ideas, and introducing multiple domains come into play to remove the Jacobian term. Using assumption 4 in Lemma1, for $\mb{u} = \mb{u_0}, \dots, \mb{u_{2n_s}}$, we have $2n_s+1$ equations like Equation~\ref{nothing}. Therefore, we can remove the effect of the Jacobian term by taking the difference for every equation $\mb{u} = \mb{u_1}, \dots,\mb{u_{2n_s}}$ with the equation where $\mb{u} = \mb{u_0}$:
\begin{equation}
    \sum_{i=1}^n (\frac{\partial^2 q_i(z_i, \mb{u_k})}{\partial z_i^2}- \frac{\partial^2 q_i(z_i, \mb{u_0})}{\partial z_i^2})\frac{\partial z_i } {\partial \hat{z}_j} \frac{\partial z_i}{\partial \hat{z}_q }  +  (\frac{\partial q_i(z_i, \mb{u_k})}{\partial z_i}- \frac{\partial q_i(z_i, \mb{u_0})}{\partial z_i})\frac{\partial ^2 z_i}{\partial \hat{z}_j \partial \hat{{z}}_q} = 0 .
\end{equation}
 For invariant variables $\mb{z}_c$, their log density doesn't change across different domains. Thus, we can get rid of invariant parts of the equation above and have
\begin{equation}
        \sum_{i=1}^{n_s} (\frac{\partial^2 q_i(z_i, \mb{u_k})}{\partial z_i^2}- \frac{\partial^2 q_i(z_i, \mb{u_0})}{\partial z_i^2})\frac{\partial z_i } {\partial \hat{z_j}} \frac{\partial z_i}{\partial \hat{z_q} }  +  (\frac{\partial q_i(z_i, \mb{u_k})}{\partial z_i}- \frac{\partial q_i(z_i, \mb{u_0})}{\partial z_i})\frac{\partial ^2 z_i}{\partial \hat{z_j} \partial \hat{{z_q}}} = 0 .
\end{equation}
Simplify the notation as \text{$\phi''_i(\mb{k}):= \frac{\partial^2 q_i(z_i, \mb{u_k})}{ \partial z_i ^2} - \frac{\partial^2 q_i(z_i, \mb{u_0)}}{ \partial z_i ^2}$ ,  $\phi'_i(\mb{k}):= \frac{\partial q_i(z_i, \mb{u_k})}{ \partial z_i } - \frac{\partial q_i(z_i, \mb{u_0)}}{ \partial z_i }$}and rewrite those equations above as a linear system, we have 

\begin{equation}
    \hspace{-1cm}
\begin{bmatrix}
\phi_1''(\mb{1}) & \hdots & \phi_{i}''(\mb{1}) & \hdots & \phi_{n_s}''(\mb{1})  & \phi_1'(\mb{1}) & \hdots & \phi_{i}'(\mb{1}) & \hdots & \phi_{n_s}'(\mb{1})\\
\vdots & \ddots & \vdots & \vdots  & \vdots  &  \vdots & \vdots & \ddots & \vdots & \vdots\\
\phi_1''(\mb{k}) & \hdots & \phi_i''(\mb{k})  & \hdots & \phi_{n_s}''(\mb{k}) &  \phi_1'(\mb{k}) & \hdots & \phi_i'(\mb{k})  & \hdots & \phi_{n_s}'(\mb{k}) \\
\vdots & \ddots & \vdots & \vdots  & \vdots  &  \vdots & \vdots & \ddots & \vdots & \vdots\\
\phi_{1}''(\mb{2n_s}) & \hdots & \phi_i''(\mb{2n_s})  & \hdots & \phi_{n_s}''(\mb{2n_s}) &  \phi_1'(\mb{2n_s}) & \hdots & \phi_i'(\mb{2n_s})  & \hdots & \phi_{n_s}'(\mb{2n_s}) \\
\end{bmatrix}
\begin{bmatrix}
    \frac{\partial z_1}{ \partial \hat{z}_j }\frac{\partial z_1}{\partial \hat{z}_q} \vspace{0.3cm} \\ 
    \vdots \vspace{0.3cm} \\
    \frac{\partial z_{n_s}}{ \partial \hat{z}_j }\frac{\partial z_{n_s}}{\partial \hat{z}_q} \vspace{0.3cm}  \\ 
    \frac{\partial^2 z_1}{\partial \hat{z}_j\hat{z}_q } \vspace{0.3cm} \\ 
    \vdots \vspace{0.3cm} \\
    \frac{\partial^2 z_{n_s}}{\partial \hat{z}_j\hat{z}_q } \vspace{0.3cm} \\ 
\end{bmatrix}
= \mb{0} .
\end{equation}

Thus, if the above matrix is invertible according to assumption 4 in Theorem~1, we will leave its null space all zero. i.e., $
\frac{\partial z_i^2}{ \partial \hat{z}_j \hat{z}_q} = 0 $ and $\frac{\partial z_i}{\partial \hat{z}_j} \frac{\partial z_i}{\partial \hat{z}_q}=0 $ for all $i \in \{1, \dots, n_s\}, j,q \in \{1,\dots,n\}, j\not=q$. We further use the property that the $h$ is invertible, which means for the Jacobian matrix of transformation $h$:
\begin{equation}
    J_h = \begin{bmatrix}
        \frac{\partial \mb{z}_c}{ \partial \mb{\hat{z}}_c} & \frac{\partial \mb{z}_c}{ \partial \mb{\hat{z}}_s} \\
        \frac{\partial \mb{z}_s}{\partial \mb{\hat{z}}_c} & \frac{\partial\mb{ z_s}}{ \partial \mb{\hat{z}}_s}
    \end{bmatrix} .
\end{equation}
the $[\frac{\partial \mb{z}_s}{\partial \mb{\hat{z}}_c} , \frac{\partial\mb{ z_s}}{ \partial \mb{\hat{z}}_s}]$ contains only one non zero value in each row. As proven in Appendix \ref{subspace}, we can get $\frac{\partial \mb{z}_s}{\partial \hat{\mb{z}}_c} = 0 $ with number of domains larger or equal to $n_s+1$. 
Thus,  $\frac{\partial\mb{ z_s}}{ \partial \mb{\hat{z}}_s}$ is an invertible full rank-matrix with only one nonzero value in each row. The changing variable $\mb{z}_s$ is component-wise identifiable.

\vspace{1cm}
\subsection{Discussion of component-wise identifiability of repeated distribution for partial changing variables} \label{appB}
In this section, we start with an example to discuss the possible scenarios where there are repeated distributions for partially changing variables among different domains. Based on this example, we proceed to provide an intuitive proof of Remark~\ref{remark1}.

Let's follow the proof of component-wise identifiability of changing variables. We directly look into the equation
\begin{equation} 
    0 = \sum_{i=1}^n \frac{\partial^2 q_i(z_i, \mb{u})}{\partial z_i^2} \frac{\partial z_i }{\partial \hat{z}_j} \frac{\partial z_i}{\partial \hat{z}_q } + \frac{\partial q_i(z_i ,\mb{u})}{\partial z_i} \frac{\partial ^2 z_i}{\partial \hat{z}_j \partial \hat{{z}}_q} + \frac{\partial^2 \log|J_{{h}}| }{\partial \hat{z}_j \partial \hat{{z}}_q} . 
\end{equation}
Our goal is to produce the matrix containing $\frac{\partial^2 q_i(z_i, \mb{u})}{\partial z_i^2} $ and $\frac{\partial q_i(z_i ,\mb{u})}{\partial z_i}$ whose null space only contains zero vector. However, we can't ensure every arrived domain will bring enough change. In this case, distributions of the same variable on different domains may be the same. i.e., $q_i(z_i, \mb{u_l}) = q_i(z_i, \mb{u_k})$ where $l \not= k $. Our discussion will mainly revolve around this situation.

Let's start with the simplest case where there are only two changing variables $z_1$ and $z_2$ and no invariant variables.  We know from Theorem1 that we need $2n_s+1$ domains to reveal their component-wise identifiability. Therefore, for $\mb{u} = \mb{u_0}, \dots , \mb{u_4}$, we have the following linear system:

\begin{equation*}
\hspace{-1.5cm}
\begin{bmatrix}
\phi_1''(\mb{1},\mb{0}) & \phi_2''(\mb{1},\mb{0})  & \phi_1'(\mb{1},\mb{0}) & \phi_2'(\mb{1},\mb{0})\\ 
\phi_1''(\mb{2},\mb{0}) & \phi_2''(\mb{2},\mb{0})  & \phi_1'(\mb{2},\mb{0}) & \phi_2'(\mb{2},\mb{0}) \\ 
\phi_1''(\mb{3},\mb{0}) & \phi_2''(\mb{3},\mb{0})  & \phi_1'(\mb{3},\mb{0}) & \phi_2'(\mb{3},\mb{0}) \\ 
\phi_1''(\mb{4},\mb{0}) & \phi_2''(\mb{4},\mb{0})  & \phi_1'(\mb{4},\mb{0}) & \phi_2'(\mb{4},\mb{0}) \\ 
\end{bmatrix}
\begin{bmatrix}
    \frac{\partial z_1}{ \partial \hat{z_1} }\frac{\partial z_1}{\partial \hat{z_2}} \vspace{0.3cm} \\ 
    \frac{\partial z_2}{ \partial \hat{z_1} }\frac{\partial z_2}{\partial \hat{z_2}} \vspace{0.3cm} \\ 
    \frac{\partial^2 z_1}{\partial \hat{z_1}\hat{z_2} } \vspace{0.3cm} \\ 
    \frac{\partial^2 z_2}{\partial \hat{z_1}\hat{z_2} } \vspace{0.3cm} \\ 
\end{bmatrix}
= \mb{0} .
\end{equation*}
where $\phi''_i(\mb{k},\mb{l}) := \frac{\partial^2 q_i(z_i, \mb{u_k})}{ \partial z_i ^2} - \frac{\partial^2 q_i(z_i, \mb{u_l)}}{ \partial z_i ^2}$ ,  $\phi'_i(\mb{k},\mb{l}) := \frac{\partial q_i(z_i, \mb{u_k})}{ \partial z_i } - \frac{\partial q_i(z_i, \mb{u_l)}}{ \partial z_i }$

Assume $z_1$ varies sufficiently across all domains. i.e., $q_1(z_1, \mb{u_j}) \not= q_1(z_1, \mb{u_k})$ for all $ k,l \in \{1,\dots, 5\}$, while $z_2$ partially changes across domains, e.g., $q_2(z_2,\mb{u_0}) \not= q_2(z_2, \mb{u_1}) = q_2(z_2, \mb{u_2}) = q_2(z_2, \mb{u_3}) =q_2(z_2, \mb{u_4})$. Subtract the first row with other rows, we have
\begin{equation*}
    \hspace{-1.5cm}
\begin{bmatrix}
\phi_1''(\mb{1},\mb{0}) & \phi_2''(\mb{1},\mb{0})  & \phi_1'(\mb{1},\mb{0}) & \phi_2'(\mb{1},\mb{0}) \\ 
\phi_1''(\mb{2},\mb{1}) & 0  & \phi_1'(\mb{2},\mb{1}) & 0 \\ 
\phi_1''(\mb{3},\mb{1}) & 0  & \phi_1'(\mb{3},\mb{1}) & 0 \\ 
\phi_1''(\mb{4},\mb{1}) & 0  & \phi_1'(\mb{4},\mb{1}) & 0 \\ 
\end{bmatrix}
\begin{bmatrix}
    \frac{\partial z_1}{ \partial \hat{z}_1 }\frac{\partial z_1}{\partial \hat{z}_2} \vspace{0.3cm} \\ 
    \frac{\partial z_2}{ \partial \hat{z}_1 }\frac{\partial z_2}{\partial \hat{z}_2} \vspace{0.3cm} \\ 
    \frac{\partial^2 z_1}{\partial \hat{z}_1\hat{z}_2 } \vspace{0.3cm} \\ 
    \frac{\partial^2 z_2}{\partial \hat{z}_1\hat{z}_2 } \vspace{0.3cm} \\ 
\end{bmatrix}
= \mb{0} .
\end{equation*}
Apparently, the matrix above is not invertible as the second column and fourth column are dependent. 
What if we further release the condition by introducing new changing domains? i.e., $q_2(z_2,\mb{u_0}) \not= q_2(z_2, \mb{u_1}) \not= q_2(z_2, \mb{u_2}) = q_2(z_2, \mb{u_3}) =q_2(z_2, \mb{u_4})$. We will have the following linear system:
\begin{equation*}
    \hspace{-1.5cm}
\begin{bmatrix}
\phi_1''(\mb{1},\mb{0}) & \phi_2''(\mb{1},\mb{0})  & \phi_1'(\mb{1},\mb{0}) & \phi_2'(\mb{1},\mb{0}) \\ 
\phi_1''(\mb{2},\mb{1}) & \phi_2''(\mb{2},\mb{1})  & \phi_1'(\mb{2},\mb{1}) & \phi_2'(\mb{2},\mb{1}) \\ 
\phi_1''(\mb{3},\mb{1}) & 0  & \phi_1'(\mb{3},\mb{1}) & 0 \\ 
\phi_1''(\mb{4},\mb{1}) & 0  & \phi_1'(\mb{4},\mb{1}) & 0 \\ 
\end{bmatrix}
\begin{bmatrix}
    \frac{\partial z_1}{ \partial \hat{z}_1 }\frac{\partial z_1}{\partial \hat{z}_2} \vspace{0.3cm} \\ 
    \frac{\partial z_2}{ \partial \hat{z}_1 }\frac{\partial z_2}{\partial \hat{z}_2} \vspace{0.3cm} \\ 
    \frac{\partial^2 z_1}{\partial \hat{z}_1\hat{z}_2 } \vspace{0.3cm} \\ 
    \frac{\partial^2 z_2}{\partial \hat{z}_1\hat{z}_2 } \vspace{0.3cm} \\ 
\end{bmatrix}
= \mb{0} .
\end{equation*}

From the example above, we can easily prove the Remark~\ref{remark1} from its contra-positive perspective, which is to prove that for $n_s \geq 2$, if $|S_i| \leq 2$, then the Lemma~\ref{lemma1} cannot hold. We denote the matrix in Lemma~1.4 as $W$ and use $W_{:,j}$ to denote the $jth$ column of the matrix $W$. For any latent changing variable $z_i$ whose $|S_i| \leq 2$, there are at most two different distributions across all observed domains. Thus, there is at most one nonzero entry in $W_{:,i}$ and $W_{:,2i}$, which will directly lead to the linear dependence between $W_{:,i}$ with $W_{:,2i}$. The Lemma~\ref{lemma1} cannot hold. We now provide some empirical results.

\begin{wrapfigure}{r}{0.5\textwidth} 
\vspace{-0.5cm}
    \centering
    \includegraphics[width=0.5\textwidth]{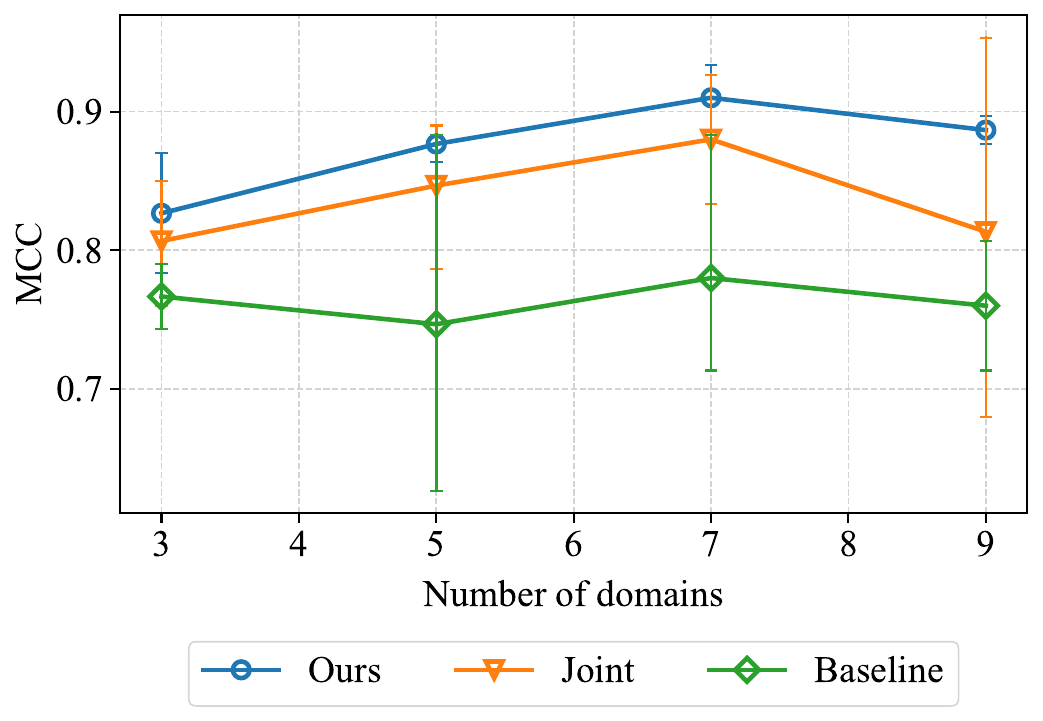} 
    \caption{Comparison of identifiability for $z_1$ using Joint training and our method qualitatively and quantitatively. }
    \label{fig:15domain}
\end{wrapfigure}
\FloatBarrier

\textbf{Repeated distributions for partial changing variables.} We conduct the experiment using data $n_s = 2, n=4$ of Gaussian. We test the case that $z_{s,1}$ have changing distributions over all distributions while $z_{s,2}$ only holds three different distributions across 
domains. As shown in Figure~\ref{fig:15domain}, our method outperforms both joint train and baseline. It may be because our method has the ability to maintain the performance learned from previous distributions and prevents potential impairment from new arriving distributions with repeated distributions as discussed in Section~\ref{sec: Ident}. For this instance, our method exhibits more robust performance than joint training against negative effects from $8,9$th distributions.

\section{Discussion of those properties} 
\subsection{Possible impairment for partial changing variables when new domains are involved}
\label{dicussion_partial}

Before we dive into the details, we need to first clarify one basic concept: \textbf{Incremental domains can't affect the overall identifiability of all changing variables theoretically}. As long as both Theorem~\ref{thm1} and Lemma~\ref{lemma1} state that the overall identifiability is determined by the distribution of true latent changing variables, the way of learning can't influence it theoretically. However, the identifiability of partial changing variables will be affected as shown in Section~3.2.2 and Experiment~4.2. We demonstrate the influence of the new domain on the identifiability of partial variables through a carefully designed example, as inferred below.
\begin{proposition}
    For latent variables $z_1, \dots, z_n$ with sequentially arriving domains $u_0, \dots, u_T$ whose generation process follows \eqref{data gen} and Lemma1.1,1.2,1.3 hold. Assume for latent variable $z_i, i\in\{1,\dots,n\}$, the first change happens at $u_{2i-1}$ and the following domains will bring sufficient change, i.e., $p(z_i|u_0)= \dots = p(z_i|u_{2i-2}) \not=p(z_i|u_{2i-1})\not=\dots\not=p(z_i,u_T)$. Then the latent variable $z_i$ reaches component-wise identifiability after observing $u_{2i+2k\leq T}$ where $k=\{0,1,\dots|2i+2k\leq T\}$, and its identifiability degrades to subspace level after the observation of $u_{2i+2k+1}$ where $k=\{0,1,\dots|2i+2k+1\leq T\}$.
\end{proposition}

\textit{Proof} As long as the distribution of latent variables $z_i$ follows the $p(z_i|u_0)= \dots = p(z_i|u_{2i-2}) \not=p(z_i|u_{2i-1})\not= \dots \not=p(z_i,u_T)$. The latent variable $z_i$ can only be referred to as "changing variable" after observation of $u_{2i-1}$. In other words, before the observation of $u_{2i-1}$, the changing variables only include $\{ z_1, \dot, z_{i-1} \}$. At this moment, we have $\{u_0, \dots, u_{2i-2} \}$ different domains with sufficient change, and the requirement of Lemma~1.4 is satisfied. The latent variables $\{z_1, \dots, z_{i-1}\}$ are component-wise identifiable. 

When the observation of $u_{2i-1}$ happens, a new changing variable $z_i$ is introduced. The condition in Lemma~1.4 doesn't hold anymore as there are only $2n_s$ domains while $2n_s+1$ are required. However, the conditions in Theorem~1 are still fulfilled, thus the identifiability of $\{z_1, \dots, z_{i-1}\}$ degrades from component-wise level to subspace level.

\begin{wrapfigure}{r}{0.3\textwidth} 
    \centering
    \vspace{-2cm}
    \includegraphics[width=0.3\textwidth]{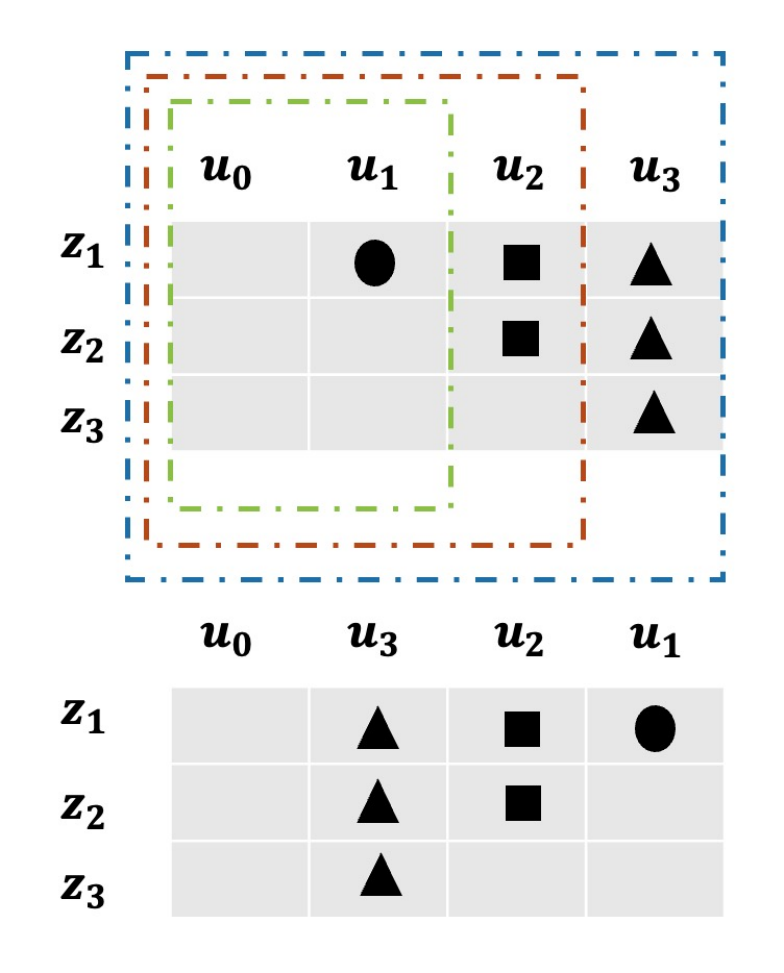} 
    \caption{\textbf{A toy example with three variables and four domains.} $z_1$ changes in $\mb{u}_1, \mb{u}_2, \mb{u}_3$, $z_2$ changes in $\mb{u}_2, \mb{u}_3$, and $z_3$ changes in $\mb{u}_3$.}
    \label{ordermatter}
\end{wrapfigure}  
\FloatBarrier

\subsection{Another property} \label{another property}

\textbf{Learning order matters.} Comparing both cases in Figure~\ref{ordermatter}, they show the same identifiability considering all domains. However, we observe that in the top case, each new domain introduces a new changing variable, while in the bottom case, the domain order is reversed. Apparently, we can achieve subspace identifiability after learning each new domain in the top case, indicating that we can progressively improve our understanding and representation of the underlying causal factors with the arrival of each new domain. However,  we can only achieve subspace identifiability until learning all domains in the bottom case. This is in line with the current learning system, where we first learn subjects with fewer changes before moving on to subjects with more 
complexity.

\newpage
\section{Pseudo Code} \label{alg}
        \begin{algorithm}[H]
\caption{Continual Nonlinear ICA}\label{continual_ica}
 \begin{algorithmic}
\Require Training data sequentially arriving $\{ \mb{x|u_1}, \dots, \mb{x|u_T} \}$
\State \text{Kaiming\_init}($\boldsymbol{\theta}$), $\mathcal{M}_t \leftarrow \lbrace\rbrace$ for all $t = 1, \ldots, T$
\For{$\mb{u} = \mb{u_1}, \dots, \mb{u_T}$}:
\For{$\{ \mb{x_1}, \dots, \mb{x_d} \} | \mb{u}$}
 \State $\mathcal{M}_t \leftarrow \mathcal{M}_t \cup \text{random select}$ $\mb{x}$ 
 \State Calculate loss $\mathcal{L}(\bs{\theta})$ as \eqref{vaeloss}
 \State $\mb{v} \leftarrow  \nabla_\mb{\theta}\mathcal{L}(\bs{\theta}, \mb{x}) $
 \State $\mb{v}_k \leftarrow   \nabla_\mb{\theta}\mathcal{L}(\bs{\theta}, \mathcal{M}_k) $ for all $k<t$
 \State $\mb{v}' \leftarrow $ Solve quadratic programming as \eqref{qp}   
 \State $\bs{\theta} \leftarrow \bs{\theta} - \alpha \mb{v}'$
\EndFor
\EndFor
\State \textbf{Return} $\bs{\theta}$

\end{algorithmic}
\end{algorithm}

\section{Visualization}\label{vis}
To provide a more intuitive demonstration of the identifiability of changing variables and compare our method with joint training, we conducted an experiment in the following setting: with $n_s = 2, n = 4$, and $\mb{z}$ values generated from a Gaussian distribution across 15 domains. We stored models trained on subsets of the training data containing 3, 5, 7, and 9 domains, a part of the whole 15 domains respectively. The test set consisted of all 15 domains, and we used these models to sample corresponding $\hat{\mb{z}}$ values. These generated $\hat{\mb{z}}$ values were then compared to the ground truth values of $\mb{z}$ for evaluation.

\begin{figure} 
    \centering
    \includegraphics[width=1.1\linewidth]{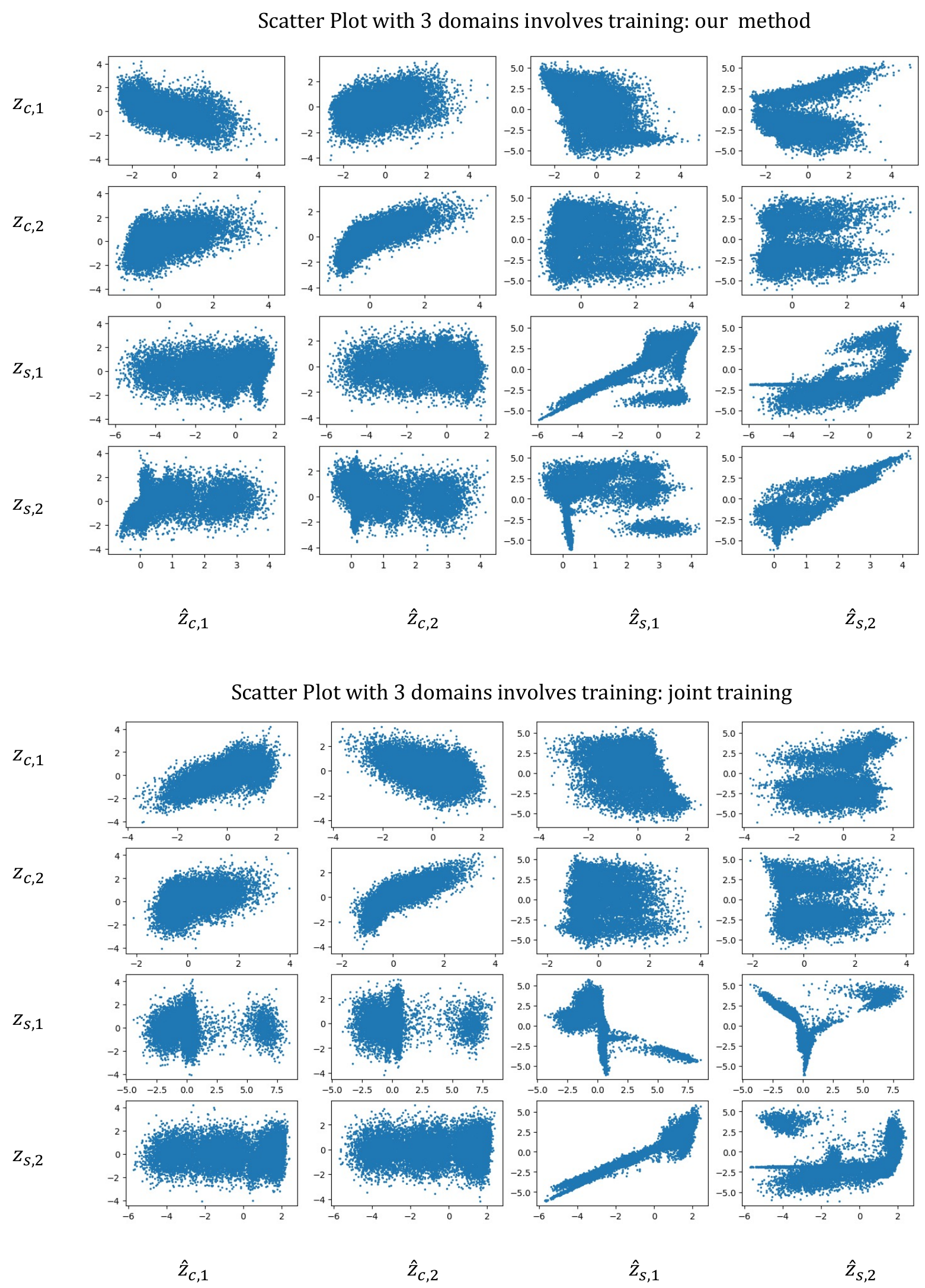}
    \caption{Visual comparison of our methods with joint training in setting that $\mb{z}$ are Gaussian, $n_s=2, n=4$. One should note that this shows the model evaluated over all 15 domains while 3 domains involve in training. }
    \label{fig:3_}
\end{figure}

\begin{figure}
    \centering
    \includegraphics[width=1.1\linewidth]{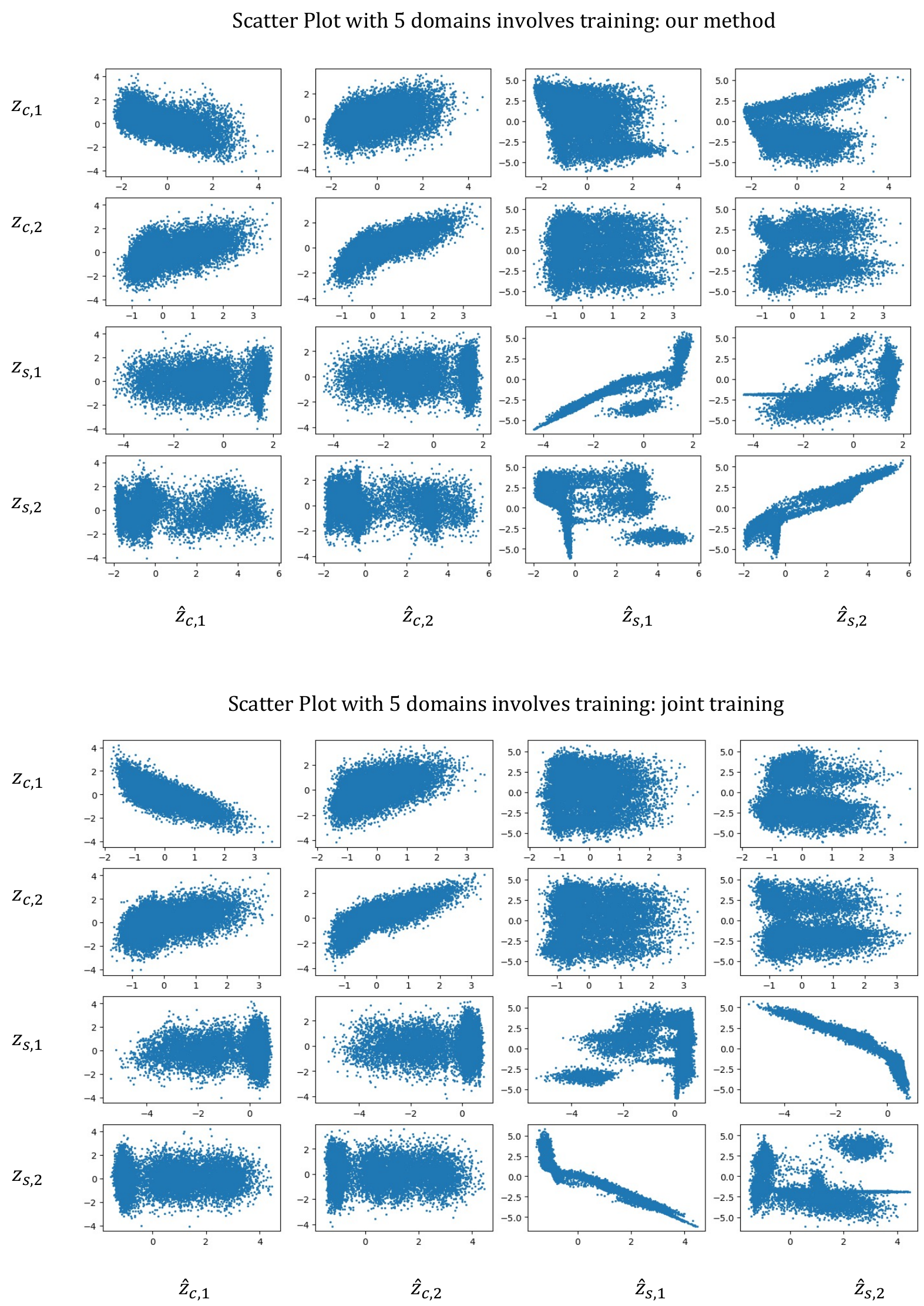}
    \caption{Visual comparison of our methods with joint training in setting that $\mb{z}$ are Gaussian, $n_s=2, n=4$. One should note that this shows the model evaluated over all 15 domains while 5 domains involve in training. }
    \label{fig:5_}
\end{figure}

\begin{figure}
    \centering
    \includegraphics[width=1.1\linewidth]{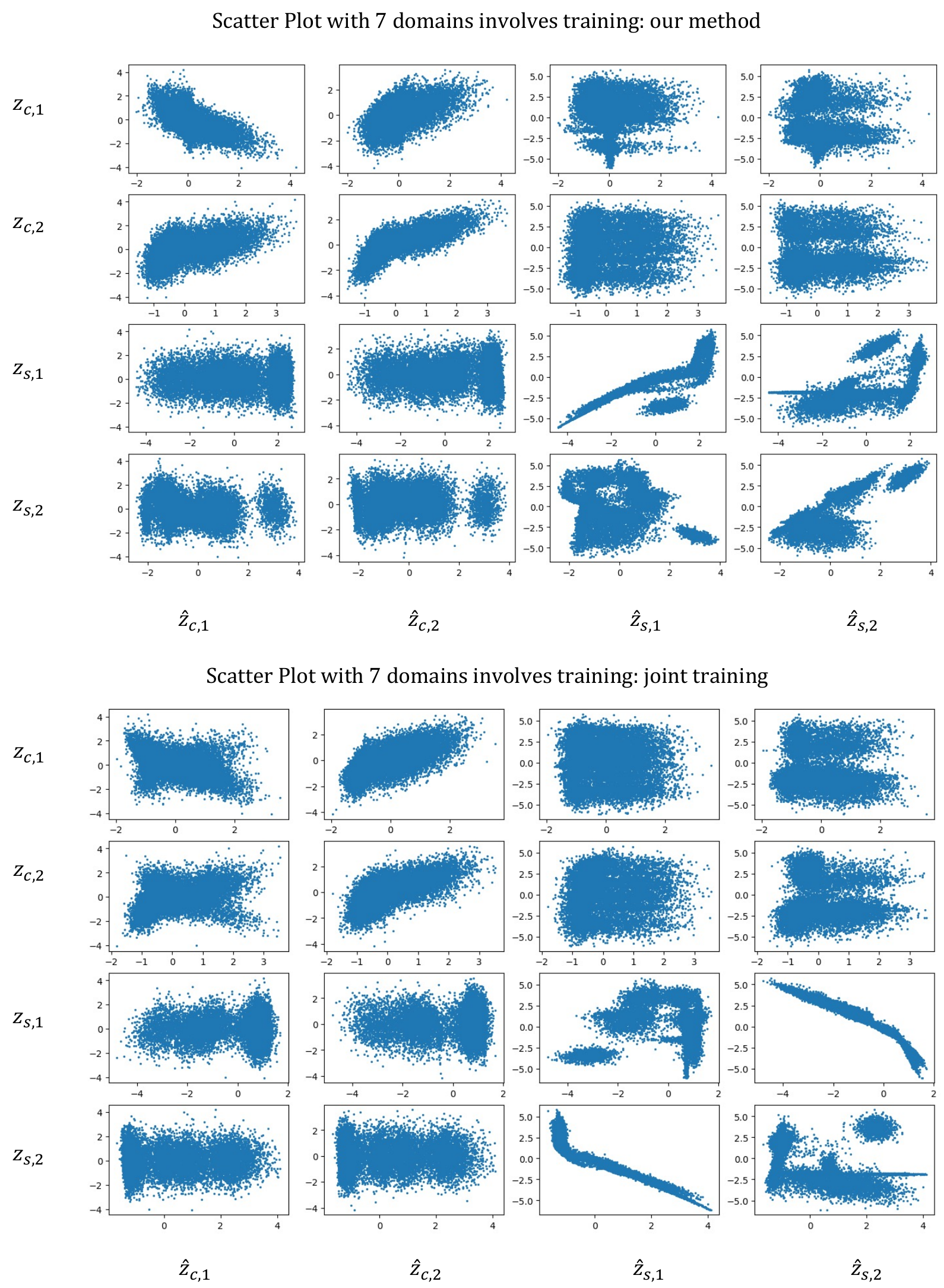}
    \caption{Visual comparison of our methods with joint training in setting that $\mb{z}$ are Gaussian, $n_s=2, n=4$. One should note that this shows the model evaluated over all 15 domains while 7 domains involve in training. }
    \label{fig:7_}
\end{figure}

\begin{figure}
    \centering
    \includegraphics[width=1.2\linewidth]{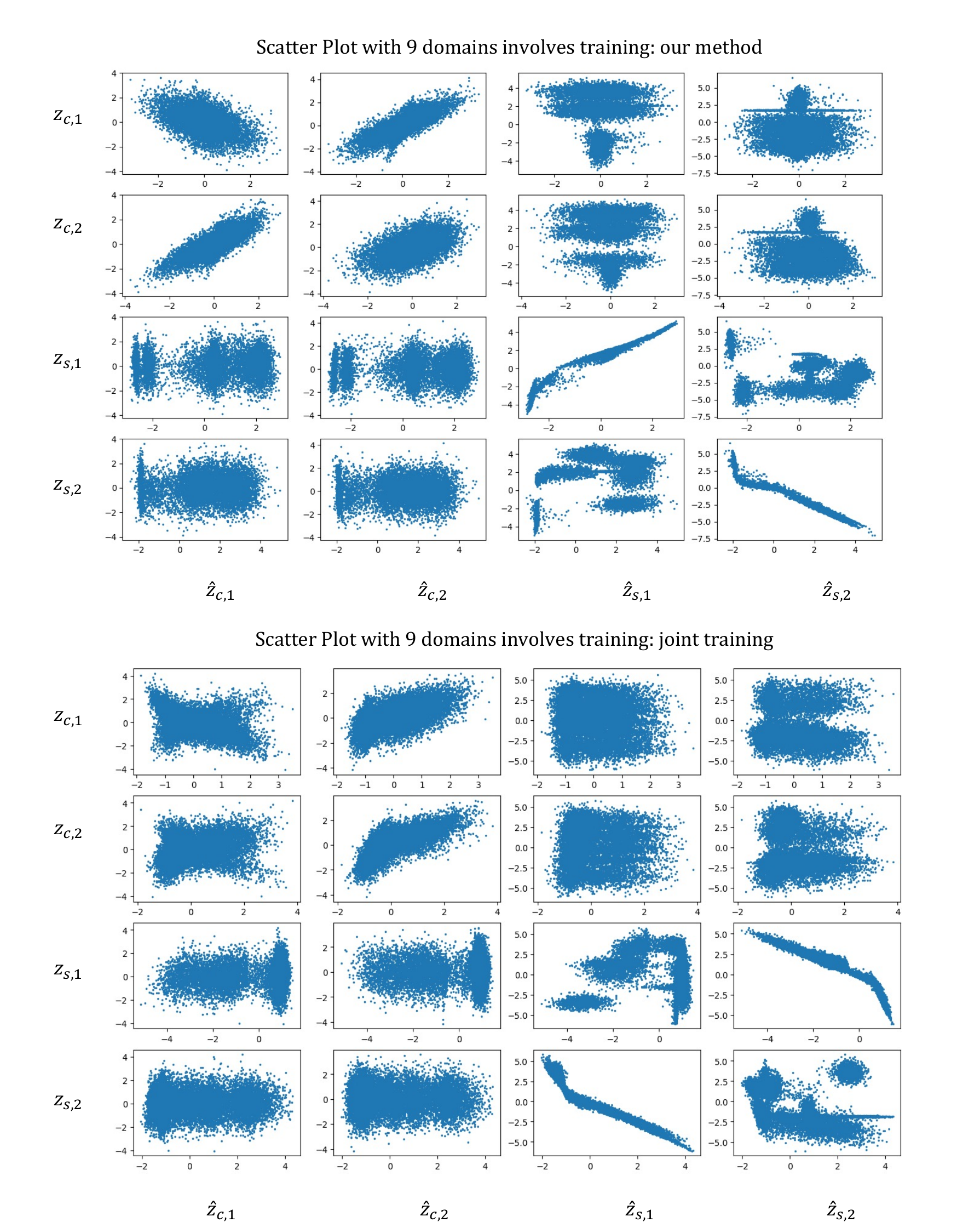}
    \caption{Visual comparison of our methods with joint training in setting that $\mb{z}$ are Gaussian, $n_s=2, n=4$. One should note that this shows the model evaluated over all 15 domains while 9 domains involve in training. }
    \label{fig:9_}
\end{figure}

Specifically, we provide the scatter plot of true latent variables $\mb{z}$ with the estimated variables $\hat{\mb{z}}$ in Figure~\ref{fig:3_},\ref{fig:5_},\ref{fig:7_},\ref{fig:9_} for both our methods and joint training. 
Figure~\ref{fig:3_},\ref{fig:5_},\ref{fig:7_},\ref{fig:9_} corresponds to a different training set that includes 3, 5, 7, and 9 domains respectively. For each figure, $\hat{\mb{z}}_{s,i}$ represents the $i$th estimated changing variable, $\hat{\mb{z}}_{c,i}$ represents $i$th estimated invariant variable, $\mb{z}_{s,i}$ represents the $i$th true changing variable and $\mb{z}_{c,i}$ represents the $i$th true invariant variable.

Based on the experiment results, we observe a stronger linear correlation that appears for estimated changing variables with real changing ones as more domains are included in the training process for both our method and joint training. That is, more domains will imply stronger identifiability, aligned with our expectations. Beyond that, our approach shows slightly inferior or even comparable performance compared to joint training, demonstrating its effectiveness.

\section{Broader Impacts}
The primary goal of our proposed method is to draw the attention to the problem of enabling the model learn identifiable representations in a sequential manner. 
 This task is essential and has broad applications. We are confident that our method will be beneficial and will not result in negative societal impacts.


\section{Experiment Details} \label{experiment design}

\subsection{Data}
We follow the data generation defined in Equation~1. Specifically, we discuss Gaussian cases where $\mb{z}_c \sim N(\mb{0}, \mb{I})$, $z_s \sim N(\mu_{\mb{u}}, \sigma_{\mb{u}}^2 \mb{I})$ for both $n_s = 2, n = 4 $ and $n_s = 4, n=8$. For each domain $\mb{u}$, the $\mu_{\mb{u}} \sim Uniform(-4,4)$ and $\sigma_{\mb{u}}^2 \sim Uniform(0.01, 1)$. 

We also discuss  the mixed Gaussian case or both $n_s = 2, n = 4 $ and $n_s = 4, n=8$ where $\mb{z}_s$ is the normalization of mixing of two Gaussian variables $N(\mb{0}, \mb{I})$ and $N(\mb{0.25}, \mb{I})$ with domain-specific modulation and translation. For $n_s=2, n=4$, each domain contains 10000 samples for training and 1000 samples for testing. For $n_s=4, n=8$, each domain contains 5000 samples for training and 1000 samples for testing. Specifically, we first mix those two Gaussian and do the normalization. After that, we modulate the normalized variable on every domain with a random variable sampled from $Uniform(0.01, 1)$. Then, we translate it with a random variable sampled from $Uniform(-4, 4)$.

\subsection{Mean correlation coefficient}
Mean correlation coefficient(MCC) is a standard metric for evaluating the recovery of latent factors in ICA literature. It averages the absolute value of the correlation coefficient between true changing variables with the estimated ones. As stated in our paper, the Lemma1 can only guarantee component-wise identifiability, leaving it unfair to directly calculate. e.g., $\mb{z} = \hat{\mb{z}}^2$ will give the correlation 0 (One should note this cannot happen in our case as $h(x) = x^2 $ is not invertible, this is just an illustrative example).

We thus use MLP to remove the nonlinearity. Specifically, we first solve a linear sum assignment problem in polynomial time on the computed correlation matrix to pair the true variables with the estimated ones. We divide these pairs into a 50-50 ratio for the training set and the test set. For each pair, we train a MLP on the training set to do the regression to remove the nonlinearity and use the test set to calculate the correlation coefficient. We then average those correlation coefficients to get the final result.

\subsection{Network architecture, hyper-parameters, Resource}
\textbf{Network Architecture. }We follow the design of \cite{kong2022partial} and summarize our specific  network architecture in Table~\ref{tab:arch-details}, which includes VAE encoder and decoder. We use component-wise spline flows \cite{durkan2019neural} to modulate the changing components $\hat{\mb{z}}_s$.

\begin{table}[ht]
\caption{Architecture details. BS: batch size, i\_dim: input dimension, z\_dim: latent dimension, s\_dim: the dimension of latent changing variables, LeakyReLU: Leaky Rectified Linear Unit.}
\label{tab:arch-details}
\resizebox{\textwidth}{!}{%
\begin{tabular*}{\textwidth}{@{\extracolsep{\fill}}|lll|}

\toprule
\textbf{Configuration} & \textbf{Description} & \textbf{Output} \\
\midrule
\textbf{1. MLP-Encoder} & Encoder for Synthetic Data & \\
\midrule
Input & Observed data from one domain $\mathbf{x}|\mathbf{u}$ & BS $\times$ i\_dim \\
Linear & 32 neurons & BS $\times$ 32 \\
Linear with Activation & 32 neurons, LeakyReLU & BS $\times$ 32 \\
Linear with Activation & 32 neurons, LeakyReLU & BS $\times$ 32 \\
Linear with Activation & 32 neurons, LeakyReLU & BS $\times$ 32 \\ 
Linear with Activation & 32 neurons, LeakyReLU & BS $\times$ 32 \\
Activation & LeakyReLU & BS $\times$ 32 \\
Linear & 32 neurons & BS $\times$ 2*z\_dim \\
\midrule
\textbf{2. MLP-Decoder} & Decoder for Synthetic Data & \\
\midrule
Input: $\hat{\mathbf{z}}$ & Sampled latent variables & BS $\times$ z\_dim \\
Linear & 32 neurons & BS $\times$ 32 \\
Linear with Activation & 32 neurons, LeakyReLU & BS $\times$ 32 \\
Linear with Activation & 32 neurons, LeakyReLU & BS $\times$ 32 \\
Linear with Activation & 32 neurons, LeakyReLU & BS $\times$ 32 \\ 
Linear with Activation & 32 neurons, LeakyReLU & BS $\times$ 32 \\
Activation & LeakyReLU & BS $\times$ 32 \\
Linear & 32 neurons & BS $\times$ i\_dim \\
\midrule
\textbf{6. Flow Model} & Flow Model to build relation between $\hat{\mathbf{z}}_s$ with $\hat{\Tilde{\mathbf{z}}}_s$ & \\
\midrule
Input & Sampled latent changing variable $\hat{\mathbf{z}}_s$ & BS $\times$ s\_dim \\
Spline flow & 8 bins, 5 bound & BS $\times$ s\_dim \\
\bottomrule
\end{tabular*}
}
\end{table}

\textbf{Training hyper-paramters} We apply Adam to train our model with 50 epochs for every domain. We use a learning rate of 0.002 with batch size of 256. We set $\alpha$ and $\beta$ both 0.1 in calculating loss as Equation~5. For the memory size of $\mathcal{M}|\mb{u}$, we randomly select 256 samples of each domain. The negative slope in Leaky-Relu is set to be 0.2.

\textbf{Resource} All experiments are run on a single Nvidia RTX A6000. In the main experiment, each seed takes 30 minutes for both Gaussian and mixed Guassian cases.

\textbf{Code} We include our code in the same zip file. For the reproduction detail, one can refer to the readme file.

\end{document}


\tableofcontents

\appendix

\subsection{Proof of component-wise identifiability of changing variables}
We divide our proof into the following parts. First, we start from the basic identifiability definition and show the true latent variables can be expressed as invertible transformations of estimated variables. We then use derivatives to construct component-wise relations between estimated variables with estimated variables. We finally show, with enough domains, we can construct the matrix whose invertibility will force the changing variable component-wise identifiable.

We start from the identifiability as introduced in : for $ \mb{u'} \in \mb{\mathcal{U}}$
\begin{equation}
        p_{\mb{x}|\mb{u}} = p_{\mb{\hat{x}}|\mb{u}} .
\end{equation}
will imply 
\begin{equation}
    p_{g(\mb{z})|\mb{u}} = p_{\hat{g}(\mb{\hat{z}})|\mb{u} .
}\end{equation}
according to the function of transformation, we can get
\begin{equation}
    p_{g^{-1} \circ g(\mb{z)|\mb{u}}} |J_{g}^{-1}| = p_{g^{-1} \circ \hat{g}(\hat{\mb{z}})|\mb{u}} |J_{g}^{-1}| .
\end{equation}
Let $\hat{h} := g^{-1} \circ \hat{g}$ to express the transformation from estimated latent variables to real latent variables, i.e., $\mb{z} = \hat{h}(\mb{\hat{z}})$. As long as both $\hat{g}$ and $g$ are invertible, the transformation $\hat{g}$ should also be invertible.  We can then get the following
\begin{equation}
    p_{\mb{z|u}} = p_{ \hat{h} (\hat{\mb{z}})|\mb{u} } {.}
\end{equation}
according to the conditional independence assumption, the log density of each marginal distribution is 
\begin{equation}
\begin{aligned}
    \log  p_{\mb{z|u }}(\mb{z}) &= \sum_{i=1}^n  \log p_{z_i|\mb{u}}(z_i) ; \\
    \log  p_{\mb{\hat{z}|u }}(\mb{z}) &= \sum_{i=1}^n  \log p_{\hat{z_i}|\mb{u}}(\hat{z_i}) .
\end{aligned}   
\end{equation}
according to the function of transformation
\begin{equation}
    p_{\mb{z|u }} = p_{(\hat{\mb{z}})|\mb{u}} |J_{{\hat{h}}^{-1}}| .
\end{equation}
Take log-density on both sides,
\begin{equation}
    \sum_{i=1}^n  \log p_{z_i|\mb{u}}(z_i) =  \sum_{i=1}^n  \log p_{\hat{z_i}|\mb{u}}(\hat{z_i}) + \log |J_{{\hat{h}}^{-1}}| .
\end{equation}
simply the notation as $q_i(z_i, \mb{u}) = \log p_{z_i|\mb{u}}(z_i), \hat{q_i}(\hat{z_i}, \mb{u}) = \log p_{z_i|\mb{u}}(z_i)$, the above equation is 
\begin{equation}
    \sum_{i=1}^n q_i(z_i, \mb{u}) =  \sum_{i=1}^n \hat{q_i}(\hat{z_i}, \mb{u}) + \log |J_{{\hat{h}}^{-1}}| . 
\end{equation}

Now we take the derivative to construct their component-wise relations. Differentiating both sides with respect to $\hat{z_j}$, $j \in [n]$, we can get
\begin{equation}
    \frac{\partial \hat{q}_j(\hat{z}_j, \mb{u})}{\partial \hat{z}_j} = \sum_{i=1}^n \frac{\partial q_i(z_i, \mb{u})}{\partial z_i} \frac{\partial z_i}{ \partial \hat{z}_j} - \frac{\partial \log |J_{{\hat{h}}^{-1}}|}{\partial \hat{z}_j} .
\end{equation}
further differentiate with respect to $\hat{z}_k$, $k \in [n], k\not=j$, according to the chain rule,
\begin{equation} \label{nothing}
    0 = \sum_{i=1}^n \frac{\partial^2 q_i(z_i, \mb{u})}{\partial z_i^2} \frac{\partial z_i }{\partial \hat{z_j}} \frac{\partial z_i}{\partial \hat{z_k} } + \frac{\partial q_i(z_i ,\mb{u})}{\partial z_i} \frac{\partial ^2 z_i}{\partial \hat{z_j} \partial \hat{{z_k}}} - \frac{\partial^2 \log|J_{{\hat{h}}^{-1}}| }{\partial \hat{z_j} \partial \hat{{z_k}}} . 
\end{equation}
This equation first time allows us to have the component-wise relation between $\mb{\hat{z}}$ with $\mb{z}$. However, the Jacobian term $\frac{\partial^2 \log|J_{{\hat{h}}^{-1}}| }{\partial \hat{z_j} \partial \hat{{z_k}}}$ is not tractable at all as we have no knowledge about $g$ (once we have, everything is solved). This is where multiple domains come into play. Following assumption 4 in Theorem1, for $\mb{u} = \mb{u_0}, \dots, \mb{u_{2n_s}}$, we have $2n_s+1$ equations like \ref{nothing}. Therefore, we can remove the effect of Jocabian term by taking the difference for every equation $\mb{u} = \mb{u_1}, \dots ,\mb{u_{2n_s}}$ with the equation where $\mb{u} = \mb{u_0}$:
\begin{equation}
    \sum_{i=1}^n (\frac{\partial^2 q_i(z_i, \mb{u_q})}{\partial z_i^2}- \frac{\partial^2 q_i(z_i, \mb{u_0})}{\partial z_i^2})\frac{\partial z_i } {\partial \hat{z_j}} \frac{\partial z_i}{\partial \hat{z_k} }  +  (\frac{\partial q_i(z_i, \mb{u_q})}{\partial z_i}- \frac{\partial q_i(z_i, \mb{u_0})}{\partial z_i})\frac{\partial ^2 z_i}{\partial \hat{z_j} \partial \hat{{z_k}}} = 0 .
\end{equation}
 For invariant variables $\mb{z}_c$, their log density don't change across different domains. Thus, we can get rid of invariant parts of equation above and have
\begin{equation}
        \sum_{i=1}^{n_s} (\frac{\partial^2 q_i(z_i, \mb{u_q})}{\partial z_i^2}- \frac{\partial^2 q_i(z_i, \mb{u_0})}{\partial z_i^2})\frac{\partial z_i } {\partial \hat{z_j}} \frac{\partial z_i}{\partial \hat{z_k} }  +  (\frac{\partial q_i(z_i, \mb{u_q})}{\partial z_i}- \frac{\partial q_i(z_i, \mb{u_0})}{\partial z_i})\frac{\partial ^2 z_i}{\partial \hat{z_j} \partial \hat{{z_k}}} = 0 .
\end{equation}
Simply the notation as $\phi''_i(\mb{q},\mb{0}) := \frac{\partial^2 q_i(z_i, \mb{u_q})}{ \partial z_i ^2} - \frac{\partial^2 q_i(z_i, \mb{u_0)}}{ \partial z_i ^2}$ ,  $\phi'_i(\mb{q},\mb{0}) := \frac{\partial q_i(z_i, \mb{u_q})}{ \partial z_i } - \frac{\partial q_i(z_i, \mb{u_0)}}{ \partial z_i }$and rewrite those equations above as a linear system, we have 
\begin{equation}
    \hspace{-2cm}
\begin{bmatrix}
\phi_1''(\mb{1},\mb{0}) & \hdots & \phi_{i}''(\mb{1},\mb{0}) & \hdots & \phi_{n_s}''(\mb{1},\mb{0})  & \phi_1'(\mb{1},\mb{0}) & \hdots & \phi_{i}'(\mb{1},\mb{0}) & \hdots & \phi_{n_s}'(\mb{1},\mb{0})\\
\vdots & \ddots & \vdots & \vdots  & \vdots  &  \vdots & \vdots & \ddots & \vdots & \vdots\\
\phi_1''(\mb{q}, \mb{0}) & \hdots & \phi_i''(\mb{q}, \mb{0})  & \hdots & \phi_{n_s}''(\mb{j},\mb{0}) &  \phi_1'(\mb{q}, \mb{0}) & \hdots & \phi_i'(\mb{q}, \mb{0})  & \hdots & \phi_{n_s}'(\mb{q},\mb{0}) \\
\vdots & \ddots & \vdots & \vdots  & \vdots  &  \vdots & \vdots & \ddots & \vdots & \vdots\\
\phi_{1}''(\mb{2n_s}, \mb{0}) & \hdots & \phi_i''(\mb{2n_s}, \mb{0})  & \hdots & \phi_{n_s}''(\mb{2n_s},\mb{0}) &  \phi_1'(\mb{2n_s}, \mb{0}) & \hdots & \phi_i'(\mb{2n_s}, \mb{0})  & \hdots & \phi_{n_s}'(\mb{2n_s},\mb{0}) \\
\end{bmatrix}
\begin{bmatrix}
    \frac{\partial z_1}{ \partial \hat{z_j} }\frac{\partial z_1}{\partial \hat{z_k}} \vspace{0.3cm} \\ 
    \vdots \vspace{0.3cm} \\
    \frac{\partial z_{n_s}}{ \partial \hat{z_j} }\frac{\partial z_{n_s}}{\partial \hat{z_k}} \vspace{0.3cm}  \\ 
    \frac{\partial^2 z_1}{\partial \hat{z_j}\hat{z_k} } \vspace{0.3cm} \\ 
    \vdots \vspace{0.3cm} \\
    \frac{\partial^2 z_{n_s}}{\partial \hat{z_j}\hat{z_k} } \vspace{0.3cm} \\ 
\end{bmatrix}
= \mb{0} .
\end{equation}

Thus, if the above matrix is invertible according to assumption 4 in Theorem, we will leave the vector all zero. i.e., $
\frac{\partial z_i^2}{ \partial \hat{z_j} \hat{z_k}} = 0 $ and $\frac{\partial z_i}{\partial \hat{z_j}} \frac{\partial z_i}{\hat{z_k}}=0 $ for all $i \in [n_s], j,k \in [n], j\not=k$. We further use the property that the $h$ is invertible, which means for the Jacobian matrix of transformation $h$
\begin{equation}
    J_h = \begin{bmatrix}
        \frac{\partial \mb{z}_c}{ \partial \mb{\hat{z}}_c} & \frac{\partial \mb{z}_c}{ \partial \mb{\hat{z}}_s} \\
        \frac{\partial \mb{z}_s}{\partial \mb{\hat{z}}_c} & \frac{\partial\mb{ z_s}}{ \partial \mb{\hat{z}}_s}
    \end{bmatrix} .
\end{equation}
the $[\frac{\partial \mb{z}_s}{\partial \mb{\hat{z}}_c} , \frac{\partial\mb{ z_s}}{ \partial \mb{\hat{z}}_s}]$ contains only one non zero value in each row.  As long as $\mb{z}_s$ change across domains while $\mb{\hat{z}}_c$ don't, the submatrix $ \frac{\partial \mb{z}_s}{\partial \mb{\hat{z}}_c}$ should all be 0. Thus, the submatrix $\frac{\partial\mb{ z_s}}{ \partial \mb{\hat{z}}_s}$ is an invertible full rank-matrix with one nonzero value in each row. The changing variable $\mb{z}_s$ are component-wise identifiable.

\subsection{Proof of component-wise identifiability of nonsufficient changing variables} \label{appB}

Let's follow the proof of component-wise identifiability of changing variables. We directly look into the equation
\begin{equation}
    \sum_{i=1}^{n_s} \left(\phi''_i(z_i, \mb{u}) \frac{\partial z_i}{\partial \hat{z_j}} \frac{\partial z_i}{\hat{z_k}} + \phi'_i(z_i, \mb{u}) \frac{\partial^2 z_i}{\partial \hat{z_j}\partial \hat{z_k}} \right) + \frac{\partial^2 \log |J_{\hat{h}} | }{\partial \hat{z_j} \partial \hat{z_k}} = 0  .
\end{equation}
where 
\begin{equation*}
\begin{aligned}
    \phi''_i(z_i, \mb{u}) &:= \frac{\partial^2 q_i(z_i,\mb{u})}{\partial \hat{z_i}^2};  \\
    \phi'_i(z_i, \mb{u}) &:= \frac{\partial q_i(z_i,\mb{u})}{\partial \hat{z_i}}  .
\end{aligned} 
\end{equation*}
Our goal is to produce a matrix containing $\phi''_i(z_i, \mb{u})$ and $\phi'_i(z_i, \mb{u})$ whose null space only contains zero vector. However, we can't ensure every arrived domain will bring enough change. In this case, distributions of the same variable on different domains may be the same. i.e., $q_i(z_i, \mb{u_j}) = q_i(z_i, \mb{u_k})$ where $j \not= k $. Our discussion will mainly revolve around this situation.

Let's start with the simplest case where there are only two changing variables $z_1$ and $z_2$.  It's well-known that we need $2n+1$ domains to reveal their identifiability. Therefore, for $\mb{u} = \mb{u_0}, \dots , \mb{u_4}$, we have the following linear system:

\begin{equation*}
\hspace{-1.5cm}
\begin{bmatrix}
\phi_1''(z_1, \mb{u_1}) - \phi_1''(z_1, \mb{u_0}) & \phi_2''(z_2, \mb{u_1}) - \phi_2''(z_2, \mb{u_0})  & \phi_1'(z_1, \mb{u_1}) - \phi_1'(z_1, \mb{u_0}) & \phi_2'(z_2, \mb{u_1}) - \phi_2'(z_2, \mb{u_0}) \vspace{0.2cm} \\ 
\phi_1''(z_1, \mb{u_2}) - \phi_1''(z_1, \mb{u_0}) & \phi_2''(z_2, \mb{u_2}) - \phi_2''(z_2, \mb{u_0})  & \phi_1'(z_1, \mb{u_2}) - \phi_1'(z_1, \mb{u_0}) & \phi_2'(z_2, \mb{u_2}) - \phi_2'(z_2, \mb{u_0})  \vspace{0.2cm} \\ 
\phi_1''(z_1, \mb{u_3}) - \phi_1''(z_1, \mb{u_0}) & \phi_2''(z_2, \mb{u_3}) - \phi_2''(z_2, \mb{u_0})  & \phi_1'(z_1, \mb{u_3}) - \phi_1'(z_1, \mb{u_0}) & \phi_2'(z_2, \mb{u_3}) - \phi_2'(z_2, \mb{u_0})  \vspace{0.2cm} \\ 
\phi_1''(z_1, \mb{u_4}) - \phi_1''(z_1, \mb{u_0}) & \phi_2''(z_2, \mb{u_4}) - \phi_2''(z_2, \mb{u_0})  & \phi_1'(z_1, \mb{u_4}) - \phi_1'(z_1, \mb{u_0}) & \phi_2'(z_2, \mb{u_4}) - \phi_2'(z_2, \mb{u_0})  \vspace{0.2cm} \\ 
\end{bmatrix}
\begin{bmatrix}
    \frac{\partial z_1}{ \partial \hat{z_1} }\frac{\partial z_1}{\partial \hat{z_2}} \vspace{0.3cm} \\ 
    \frac{\partial z_2}{ \partial \hat{z_1} }\frac{\partial z_2}{\partial \hat{z_2}} \vspace{0.3cm} \\ 
    \frac{\partial^2 z_1}{\partial \hat{z_1}\hat{z_2} } \vspace{0.3cm} \\ 
    \frac{\partial^2 z_2}{\partial \hat{z_1}\hat{z_2} } \vspace{0.3cm} \\ 
\end{bmatrix}
= \mb{0} .
\end{equation*}
Assume $z_1$ varies sufficiently across all domains. i.e., $q_1(z_1, \mb{u_j}) \not= q_1(z_1, \mb{u_k})$ for all $j, k \in [5]$, while $z_2$ partially changes across domains, e.g., $q_2(z_2,\mb{u_0}) \not= q_2(z_2, \mb{u_1}) = q_2(z_2, \mb{u_2}) = q_2(z_2, \mb{u_3}) =q_2(z_2, \mb{u_4})$. Subtract the first row with other rows, we have
\begin{equation*}
    \hspace{-1.5cm}
\begin{bmatrix}
\phi_1''(z_1, \mb{u_1}) - \phi_1''(z_1, \mb{u_0}) & \phi_2''(z_2, \mb{u_1}) - \phi_2''(z_2, \mb{u_0})  & \phi_1'(z_1, \mb{u_1}) - \phi_1'(z_1, \mb{u_0}) & \phi_2'(z_2, \mb{u_1}) - \phi_2'(z_2, \mb{u_0}) \vspace{0.2cm} \\ 
\phi_1''(z_1, \mb{u_2}) - \phi_1''(z_1, \mb{u_1}) & 0  & \phi_1'(z_1, \mb{u_2}) - \phi_1'(z_1, \mb{u_1}) & 0 \vspace{0.2cm} \\ 
\phi_1''(z_1, \mb{u_3}) - \phi_1''(z_1, \mb{u_1}) & 0  & \phi_1'(z_1, \mb{u_3}) - \phi_1'(z_1, \mb{u_1}) & 0  \vspace{0.2cm} \\ 
\phi_1''(z_1, \mb{u_4}) - \phi_1''(z_1, \mb{u_1}) & 0  & \phi_1'(z_1, \mb{u_4}) - \phi_1'(z_1, \mb{u_1}) & 0  \vspace{0.2cm} \\ 
\end{bmatrix}
\begin{bmatrix}
    \frac{\partial z_1}{ \partial \hat{z_1} }\frac{\partial z_1}{\partial \hat{z_2}} \vspace{0.3cm} \\ 
    \frac{\partial z_2}{ \partial \hat{z_1} }\frac{\partial z_2}{\partial \hat{z_2}} \vspace{0.3cm} \\ 
    \frac{\partial^2 z_1}{\partial \hat{z_1}\hat{z_2} } \vspace{0.3cm} \\ 
    \frac{\partial^2 z_2}{\partial \hat{z_1}\hat{z_2} } \vspace{0.3cm} \\ 
\end{bmatrix}
= \mb{0} .
\end{equation*}
Apparently, the matrix above is not invertible as the second column and fourth column are dependent. What if we further release the condition by introducing new changing domains? i.e., $q_2(z_2,\mb{u_0}) \not= q_2(z_2, \mb{u_1}) \not= q_2(z_2, \mb{u_2}) = q_2(z_2, \mb{u_3}) =q_2(z_2, \mb{u_4})$. We will have the following linear system:
\begin{equation*}
    \hspace{-1.5cm}
\begin{bmatrix}
\phi_1''(z_1, \mb{u_1}) - \phi_1''(z_1, \mb{u_0}) & \phi_2''(z_2, \mb{u_1}) - \phi_2''(z_2, \mb{u_0})  & \phi_1'(z_1, \mb{u_1}) - \phi_1'(z_1, \mb{u_0}) & \phi_2'(z_2, \mb{u_1}) - \phi_2'(z_2, \mb{u_0}) \vspace{0.2cm} \\ 
\phi_1''(z_1, \mb{u_2}) - \phi_1''(z_1, \mb{u_1}) & \phi_2''(z_2, \mb{u_2}) - \phi_2''(z_2, \mb{u_1})  & \phi_1'(z_1, \mb{u_2}) - \phi_1'(z_1, \mb{u_1}) & \phi_2'(z_2, \mb{u_2}) - \phi_2'(z_2, \mb{u_1}) \vspace{0.2cm} \\ 
\phi_1''(z_1, \mb{u_3}) - \phi_1''(z_1, \mb{u_1}) & 0  & \phi_1'(z_1, \mb{u_3}) - \phi_1'(z_1, \mb{u_1}) & 0  \vspace{0.2cm} \\ 
\phi_1''(z_1, \mb{u_4}) - \phi_1''(z_1, \mb{u_1}) & 0  & \phi_1'(z_1, \mb{u_4}) - \phi_1'(z_1, \mb{u_1}) & 0  \vspace{0.2cm} \\ 
\end{bmatrix}
\begin{bmatrix}
    \frac{\partial z_1}{ \partial \hat{z_1} }\frac{\partial z_1}{\partial \hat{z_2}} \vspace{0.3cm} \\ 
    \frac{\partial z_2}{ \partial \hat{z_1} }\frac{\partial z_2}{\partial \hat{z_2}} \vspace{0.3cm} \\ 
    \frac{\partial^2 z_1}{\partial \hat{z_1}\hat{z_2} } \vspace{0.3cm} \\ 
    \frac{\partial^2 z_2}{\partial \hat{z_1}\hat{z_2} } \vspace{0.3cm} \\ 
\end{bmatrix}
= \mb{0} .
\end{equation*}
If we still hold the linear independence assumption, we can still reveal its component-wise identifiability.  This can be viewed as a special case of original assumption. However, it's meaningful in the continual learning setting: we don't need every variable changes sufficiently in every arrived domain. If we extrapolate the matrix above to $n=3$,
\begin{equation*}
    \hspace{-1cm}
\begin{bmatrix}
\phi_1''(\mb{1},\mb{0}) & \phi_2''(\mb{1},\mb{0}) & \phi_3''(\mb{1},\mb{0}) & \phi_1'(\mb{1},\mb{0}) & \phi_2'(\mb{1},\mb{0}) & \phi_3'(\mb{1},\mb{0}) \\
\phi_1''(\mb{2},\mb{1}) & \phi_2''(\mb{2},\mb{1}) & \phi_3''(\mb{2},\mb{1}) & \phi_1'(\mb{2},\mb{1}) & \phi_2'(\mb{2},\mb{1}) & \phi_3'(\mb{2},\mb{1}) \\
\phi_1''(\mb{3},\mb{1}) & \phi_2''(\mb{3},\mb{1}) & 0  & \phi_1'(\mb{3},\mb{1}) & \phi_2'(\mb{3},\mb{1}) & 0 \\
\phi_1''(\mb{4},\mb{1}) & \phi_2''(\mb{4},\mb{1}) & 0  & \phi_1'(\mb{4},\mb{1}) & \phi_2'(\mb{4},\mb{1}) & 0 \\
\phi_1''(\mb{5},\mb{1}) & 0 & 0  & \phi_1'(\mb{5},\mb{1}) & 0 & 0 \\
\phi_1''(\mb{6},\mb{1}) & 0 & 0  & \phi_1'(\mb{6},\mb{1}) & 0 & 0 \\

\end{bmatrix}
\begin{bmatrix}
    \frac{\partial z_1}{ \partial \hat{z_j} }\frac{\partial z_1}{\partial \hat{z_k}} \vspace{0.3cm} \\ 
    \frac{\partial z_2}{ \partial \hat{z_j} }\frac{\partial z_2}{\partial \hat{z_k}} \vspace{0.3cm} \\ 
    \frac{\partial z_3}{ \partial \hat{z_j} }\frac{\partial z_3}{\partial \hat{z_k}} \vspace{0.3cm} \\ 
    \frac{\partial^2 z_1}{\partial \hat{z_j}\hat{z_k} } \vspace{0.3cm} \\ 
    \frac{\partial^2 z_2}{\partial \hat{z_j}\hat{z_k} } \vspace{0.3cm} \\ 
    \frac{\partial^2 z_3}{\partial \hat{z_j}\hat{z_k} } \vspace{0.3cm} \\ 
\end{bmatrix}
= \mb{0} ,
\end{equation*}
where $\phi_i''(\mb{j}, \mb{k})$ expresses $\phi_i''(z_i, \mb{u_j)} - \phi_i''(z_i, \mb{u_k)}$. If we further release the condition, we can get
\begin{equation*}
    \hspace{-1cm}
\begin{bmatrix}
0 & 0 & \phi_3''(\mb{1},\mb{0}) & 0 & 0 & \phi_3'(\mb{1},\mb{0}) \\
0 & 0 & \phi_3''(\mb{2},\mb{1}) & 0  & 0  & \phi_3'(\mb{2},\mb{1}) \\
0 & \phi_2''(\mb{3},\mb{1}) & 0  & 0  & \phi_2'(\mb{3},\mb{1}) & 0 \\
0 & \phi_2''(\mb{4},\mb{1}) & 0  & 0 & \phi_2'(\mb{4},\mb{1}) & 0 \\
\phi_1''(\mb{5},\mb{1}) & 0 & 0  & \phi_1'(\mb{5},\mb{1}) & 0 & 0 \\
\phi_1''(\mb{6},\mb{1}) & 0 & 0  & \phi_1'(\mb{6},\mb{1}) & 0 & 0 \\

\end{bmatrix}
\begin{bmatrix}
    \frac{\partial z_1}{ \partial \hat{z_j} }\frac{\partial z_1}{\partial \hat{z_k}} \vspace{0.3cm} \\ 
    \frac{\partial z_2}{ \partial \hat{z_j} }\frac{\partial z_2}{\partial \hat{z_k}} \vspace{0.3cm} \\ 
    \frac{\partial z_3}{ \partial \hat{z_j} }\frac{\partial z_3}{\partial \hat{z_k}} \vspace{0.3cm} \\ 
    \frac{\partial^2 z_1}{\partial \hat{z_j}\hat{z_k} } \vspace{0.3cm} \\ 
    \frac{\partial^2 z_2}{\partial \hat{z_j}\hat{z_k} } \vspace{0.3cm} \\ 
    \frac{\partial^2 z_3}{\partial \hat{z_j}\hat{z_k} } \vspace{0.3cm} \\ 
\end{bmatrix}
= \mb{0} .
\end{equation*}
We can easily extend into a general and special case, where every row of the matrix above has only two non-zero entries and the matrix is "double staircase-like". Under this condition, the matrix is still identifiable, and none of the assumptions in Theorem1 has been violated. 
We may want to further prove the least required domains for each variable are $3$. This can be taken as future work.











\subsection{Subspace identifiability with fewer domains}
We start with the equation that 
\begin{equation}
    \sum_{i=1}^n q_i(z_i, \mb{u}) + \log|J_h| = \sum_{i=1}^n \hat{q}_i (\hat{z}_i, \mb{u})
\end{equation}
Take the derivative with estimated invariant variable $\hat{z}_j$ where $j \in \{1, \dots, n_c \}$. We can get 
\begin{equation}
    \sum_{i=1}^n \frac{\partial q_i(z_i, \mb{u})}{\partial z_i} \frac{\partial z_i}{\partial \hat{z}_j} + 
    \frac{\partial \log |J_h|}{\partial \hat{z}_j} = \frac{\partial \hat{q}_j (\hat{z}_j, \mb{u})}{\partial \hat{z}_j}
\end{equation}
If we have multiple domains $\mb{u} = \mb{u}_0, \dots ,  \mb{u}_{n_s}$, we have $n_s+1$ equations like equation above. We can remove the intractable Jacobian by taking the difference for every equation $\mb{u} = \mb{u}_1, \dots , \mb{u}_{n_s}$ with the equation where $\mb{u} = \mb{u_0}$:
\begin{equation}
    \sum_{i=1}^n (\frac{\partial q_i(z_i, \mb{u}_q)}{\partial z_i} - \frac{\partial q_i(z_i, \mb{u}_0)}{\partial z_i} ) \frac{\partial z_i} {\partial \hat{z}_j} 
    = 
    \frac{\partial \hat{q}_j (\hat{z}_j, \mb{u}_q)}{\partial \hat{z}_j} - 
    \frac{\partial \hat{q}_j (\hat{z}_j, \mb{u}_0)}{\partial \hat{z}_j}
\end{equation}
The distribution estimated variable $\hat{z}_j$ doesn't change across all domains. Thus,
\begin{equation}
    \sum_{i=1}^n (\frac{\partial q_i(z_i, \mb{u}_q)}{\partial z_i} - \frac{\partial q_i(z_i, \mb{u}_0)}{\partial z_i} ) \frac{\partial z_i} {\partial \hat{z}_j} 
    = 0
\end{equation}
Similarly, $q_i(z_i,\mb{u})$ remains the same for $i \in \{1, \dots, n_c\}$
\begin{equation}
    \sum_{i=1}^{n_s} (\frac{\partial q_i(z_i, \mb{u}_q)}{\partial z_i} - \frac{\partial q_i(z_i, \mb{u}_0)}{\partial z_i} ) \frac{\partial z_i} {\partial \hat{z}_j} 
    = 0
\end{equation}
Thus, we can have the linear system:
\begin{equation}
    \begin{bmatrix}
        \frac{\partial q_1(z_1, \mb{u}_1)}{\partial z_1} -         \frac{\partial q_1(z_1, \mb{u}_0)}{\partial z_1} & \hdots &         \frac{\partial q_{n_s}(z_{n_s}, \mb{u}_1)}{\partial z_{n_s}} - 
        \frac{\partial q_{n_s}(z_{n_s}, \mb{u}_0)}{\partial z_{n_s}} \\
        \vdots & \hdots & \vdots \\
        
        \frac{\partial q_1(z_1, \mb{u}_{n_s})}{\partial z_1} -         \frac{\partial q_1(z_1, \mb{u}_0)}{\partial z_1} & \hdots &         \frac{\partial q_{n_s}(z_{n_s}, \mb{u}_{n_s})}{\partial z_{n_s}} - 
        \frac{\partial q_{n_s}(z_{n_s}, \mb{u}_0)}{\partial z_{n_s}}        
    \end{bmatrix}
    \begin{bmatrix}
        \frac{\partial z_1}{\partial \hat{z}_j} \\
        \vdots \\
        \frac{\partial z_{n_s}}{\partial \hat{z}_j}
    \end{bmatrix} = 0
\end{equation}
If the matrix above is invertible, its null space will only contain all zeros. Thus, 
$\frac{\partial z_i}{\partial \hat{z}_j} = 0$ for any $i \in \{1,\dots, n_s\}, j \in \{1,\dots,n_c\}$. Thus, there is no relation between the true changing variable $z_s$ with estimated invariant variable $\hat{z}_c$. All information of real changing variables $z_s$ are contained in $\hat{z_s}$.

We further look back into the Jacobian matrix
\begin{equation}
J_h = 
    \begin{bmatrix}
\frac{\partial z_c}{ \partial \hat{z}_c} & \frac{\partial z_c}{\partial \hat{z}_s} \\
\frac{\partial z_s}{\partial \hat{z}_c} &
\frac{\partial z_s}{\partial \hat{z}_s}    
    \end{bmatrix}
\end{equation}
We know that $\frac{\partial z_s}{\partial \hat{z}_c} = 0$, which is trivial intuitionally as $z_s$ is changing while $\hat{z}_c$ remains the same distribution. However, as the Jacobian matrix $J_h$ is invertible, we can utilize its property that the inverse of a block matrix is 
\begin{equation}
    \begin{bmatrix}
A & B \\
C & D \\
\end{bmatrix}^{-1}
=
\begin{bmatrix}
(A-BD^{-1}C)^{-1} & -(A-BD^{-1}C)^{-1}BD^{-1} \\
-D^{-1}C(A-BD^{-1}C)^{-1} & D^{-1} + D^{-1}C(A-BD^{-1}C)^{-1}BD^{-1} \\
\end{bmatrix}
\end{equation}
Thus, for the inverse of the Jacobian matrix above 
\begin{equation}
    {J_h}^{-1} = 
        \begin{bmatrix}
\frac{\partial \hat{z}_c}{ \partial z_c} & \frac{\partial \hat{z}_c}{\partial z_s} \\
\frac{\partial \hat{z}_s}{\partial z_c} &
\frac{\partial \hat{z}_s}{\partial z_s}    
    \end{bmatrix}
\end{equation}
The bottom left term $\frac{\partial \hat{z}_s}{\partial z_c}$ must be zero. There is no relation between the estimated changing variable $\hat{z}_s$ with true invariant variable $z_c$. The changing variable $z_s$ is subspace identifiable.